\documentclass[lettersize,journal]{IEEEtran}

\usepackage[english]{babel} 
\usepackage[T1]{fontenc}
\usepackage[utf8]{inputenc} %
\usepackage{textcomp} %

\usepackage{amssymb} %
\usepackage{amsmath}
\usepackage{amsthm}
\usepackage{algorithm2e}
\usepackage{optidef}
\usepackage[printonlyused]{acronym}
\usepackage{mathdots}
\usepackage{color}
\usepackage{enumitem}
\usepackage{pgfplots}
\pgfplotsset{compat=newest}
\usetikzlibrary{decorations.pathreplacing, positioning, calc, arrows, quotes}
\usepackage{multirow}
\usepackage{subfig}
\usepackage{relsize}
\usepackage{booktabs}
\usepackage{adjustbox}
\usepackage[pages=all,color=black,placement=top,scale=1.0,opacity=1,vshift=-0.05cm]{background}

\backgroundsetup{
	pages=all,
	color=black,
	placement=top,
	scale=1.0,
	opacity=1,
	vshift=-0.05cm,
	contents={
		\begin{minipage}{2\linewidth}
			\centering
			\scriptsize
			This article has been accepted for publication in IEEE Transactions on Geoscience and Remote Sensing. This is the author's version which has not been fully edited and
			content may change prior to final publication. Citation information: DOI 10.1109/TGRS.2022.3227061\\[240ex]
			\copyright{} 2022 IEEE. Personal use is permitted, but republication/redistribution requires IEEE permission.
			See \url{https://www.ieee.org/publications/rights/index.html} for more information.
		\end{minipage}
	}
}

\newcounter{MYtempeqncnt}

\usepackage{hyperref}
\usepackage[capitalise]{cleveref}

\usepackage{colortbl}
\definecolor{std-color}{RGB}{236 237 237}
\definecolor{highest-max-color}{RGB}{221 235 206}

\DeclareMathOperator*{\argmax}{arg\,max}

\newcommand{\R}{\mathbb{R}}
\newcommand{\N}{\mathbb{N}}
\newcommand{\K}{\mathbb{K}}
\newcommand{\OO}{\mathcal{O}}
\newcommand{\zach}{\text{Zach}}
\newcommand{\dx}{\;\mathrm{d}x}
\newcommand{\Per}{\text{Per}}
\newcommand{\ContDom}{\Omega}
\newcommand{\DiscDom}{\mathcal{N}}

\newcommand{\CovReg}[1]{\Sigma_{\epsilon, #1}}
\newcommand{\DiagReg}[1]{D_{\epsilon, #1}}
\newcommand{\LevelSetFuncd}{u}
\newcommand{\DualVard}{p}
\newcommand{\Projection}{P}
\newcommand{\ProximalMapping}{\text{prox}}
\newcommand{\tv}[1]{\vert #1 \vert_{\text{BV}}}

\graphicspath{ {./input/} }

\acrodef{3dfeats}[FL]{3DCAE features}
\acrodef{3dcae}[3D-CAE]{3D convolutional autoencoder}
\acrodef{aviris}[AVIRIS]{Airborne visible/infrared imaging spectrometer}
\acrodef{aa}[AA]{average class accuracy}
\acrodef{bgm}[BGM]{Bayesian Gaussian mixture model}
\acrodef{bv}[BV]{bounded variation}
\acrodef{gmm}[GMM]{Gaussian mixture model}
\acrodef{hsi}[HSI]{hyperspectral imaging}
\acrodef{ip}[IP]{Indian Pines}
\acrodef{kc}[KC]{kappa coefficient}
\acrodef{ksc}[KSC]{Kennedy Space Center}
\acrodef{mnf}[MNF]{Minimum Noise Fraction}
\acrodef{ms}[MS]{Mumford-Shah}
\acrodef{oa}[OA]{overall accuracy}
\acrodef{pu}[PU]{Pavia University}
\acrodef{pca}[PCA]{Principal Components Analysis}
\acrodef{pdhg}[PDHG]{primal-dual hybrid gradient}
\acrodef{sa}[SA]{Salinas}
\acrodef{snr}[SNR]{signal-to-noise ratio}
\acrodef{svm}[SVM]{support vector machine}
\acrodef{tv}[TV]{total variation}

\title{A distribution-dependent Mumford-Shah model for unsupervised hyperspectral image segmentation}

\author{Jan-Christopher Cohrs, Chandrajit Bajaj, \IEEEmembership{Fellow, IEEE}, Benjamin Berkels
\thanks{Manuscript received March 29, 2022; revised XXXX XX, 2022.}
\thanks{Jan-Christopher Cohrs and Benjamin Berkels are with the Aachen Institute for Advanced Study in Computational Engineering Science (AICES), RWTH Aachen University, Germany (email: \href{mailto:cohrs@aices.rwth-aachen.de}{cohrs@aices.rwth-aachen.de}; \href{mailto:berkels@aices.rwth-aachen.de}{berkels@aices.rwth-aachen.de}).}
\thanks{Chandrajit Bajaj is with the Oden Institute for Computational and Engineering Sciences, University of Texas at Austin, USA (email: \href{mailto:bajaj@oden.utexas.edu}{bajaj@oden.utexas.edu}).}
}%

\begin{document}

\markboth{IEEE Transactions on Geoscience and Remote Sensing,~Vol.~XX, No.~X, XXXX~2022}%
{Cohrs \MakeLowercase{\textit{et al.}}: A distribution-dependent Mumford-Shah model for unsupervised hyperspectral image segmentation}

\IEEEpubid{0000--0000/00\$00.00~\copyright~2022 IEEE}

\maketitle

\begin{abstract}
Hyperspectral images provide a rich representation of the underlying spectrum for each pixel, allowing for a pixel-wise classification/segmentation into different classes.
As the acquisition of labeled training data is very time-consuming, unsupervised methods become crucial in hyperspectral image analysis.
The spectral variability and noise in hyperspectral data make this task very challenging and define special requirements for such methods.

Here, we present a novel unsupervised hyperspectral segmentation framework.
It starts with a denoising and dimensionality reduction step by the well-established \ac{mnf} transform.
Then, the \ac{ms} segmentation functional is applied to segment the data.
We equipped the MS functional with a novel robust distribution-dependent indicator function designed to handle the characteristic challenges of hyperspectral data.
To optimize our objective function with respect to the parameters for which no closed form solution is available, we propose an efficient fixed point iteration scheme.
Numerical experiments on four public benchmark datasets show that our method produces competitive results, which outperform three state-of-the-art methods substantially on three of these datasets.
\end{abstract}


\acresetall

\section{Introduction}\label{sec:introduction}
\IEEEPARstart{I}{n} many research areas that are concerned with an image-based analysis of specimens or scenes, hyperspectral images are becoming more and more popular.
Modern \ac{hsi} sensors sample the spectral space with a very high spectral resolution, resulting in very detailed and rich representations of the spectra at each pixel \cite{KeSc07}.
\ac{hsi} is often applied in the field of remote sensing, where it is used in agriculture, mineralogy, surveillance and environmental management \cite{PaHaPl19}. The sensors consider a large part of the electromagnetic spectrum. For instance, in the case of an \ac{aviris} the visible and infrared wavelengths are sampled with a very small step size \cite{KeSc07}.
However, \ac{hsi} is also attracting more attention in areas like, e.g., biomedical imaging in the form of Fourier transform infrared spectrometry, which measures a large part of the infrared regime \cite{GrHa07}.

\ac{hsi} assumes that constituents, like materials, that are present in a scene or specimen reflect specific wavelengths, which can be seen as \emph{spectral signatures} of these constituents \cite{KeSc07}.
In particular, this allows for a pixel-wise classification of an image into different constituent classes if the spatial resolution is sufficiently small.
Many spectrometers used in remote sensing do not fulfill this requirement as their spatial resolution is rather large.
Consequently, \emph{mixed pixels} are recorded that show a mixture of spectral signatures \cite{KeSc07}, which makes it harder to distinguish between the contributing constituents.
\IEEEpubidadjcol
In fact, \emph{hyperspectral unmixing} focusses on this aspect of hyperspectral data and models each pixel as a mixture of a fixed number of discrete spectral signatures, called endmembers, with the goal to determine representations of the endmembers as well as the abundances \cite{BiPlDo12}. %
Moreover, hyperspectral data often suffers from noise, which impedes the classification of the pixels further.
The most challenging obstacle in \ac{hsi} analysis, however, is the \emph{spectral variability}.
This phenomenon describes that within a constituent class the spectra show variations caused by changes in illumination conditions and material surfaces \cite{CaCoAr19}. %
\begin{figure}
    \centering
	\input{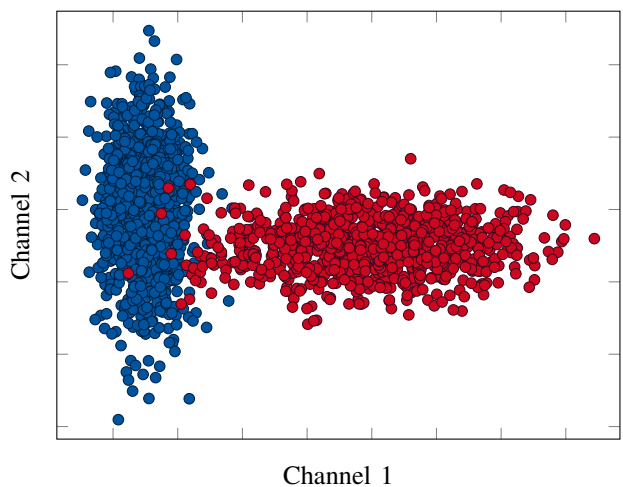}
    \caption{Toy example of an image with two segments and two channels showing strong \emph{spectral variability}.
	The axes of the plot describe the intensities of channel 1 and channel 2, respectively.
	Each dot corresponds to a spectrum.
	The spectra were drawn from two different distributions and are shown in blue and red, respectively.
	Each distribution (or cluster) represents the recorded spectra that are variations of the spectral signature of one specific constituent.
	One can see that for both clusters the deviation around the mean spectrum differs in different directions.
	Although the toy example shows clusters that follow a Gaussian distribution by design, the clusters can follow any distribution in general.}
    \label{fig:feature-variation}
\end{figure}
\cref{fig:feature-variation} shows a toy example of the spectral variability.
Segmentation and classification methods have to take this effect into account when separating the clusters and classifying the pixels.

The problem of classifying pixels into segments or classes in the hyperspectral context is mainly tackled using supervised learning.
Especially in the remote sensing community, this is a very active topic.
There is a vast amount of literature on supervised approaches to classify hyperspectral images on a per pixel basis.
Approaches that merely exploit spectral information are often based on \acp{svm} \cite{GaLiKh15,ChGl16}.
Recent spectral-spatial approaches, i.e., combining spectral with spatial information, achieve generally higher scores than spectral approaches, because they exploit more information about the scene.
Many of these recent methods are based on neural networks \cite{RoKrDu20,RoMaSo21,XuZhLi21}.
A serious shortcoming of supervised methods is the shortage of labeled data in the context of \ac{hsi} that is needed to train the methods and obtain good generalization performance.
In particular, current approaches typically train on a subset of the labeled pixels available for one image and test on the remaining labeled pixels of the very same image.
Hence, a generalization of these approaches to new data cannot be expected.
Generating labeled training data is very time-consuming.
Furthermore, because only a limited number of training samples in a very high dimensional spectral space is available, the \emph{Hughes phenomenon} \cite{Hu68} makes the classification more difficult.
Unsupervised methods avoid these problems as they do not make use of any labeled data.
The only prior knowledge that is incorporated is the number of segments that are sought.
Some approaches even try to estimate that.
Richards and Jia differentiate between \emph{a priori} labeling in the supervised and \emph{a posteriori} labeling in the unsupervised setting \cite{RiJi06}. %
The difference is that in the former case one needs a number of labeled training samples that have to be manually classified in advance, while in the latter case it is enough to take one sample from each class of interest after segmentation and assign a semantic label to the whole class with the support of spatial structures that the segmentation result provides.
A posteriori labeling even allows for semantically labeling only those segments that are of interest, which reduces the need of manual input further.

There are only few works on unsupervised hyperspectral segmentation.
Some of these focus on finding adaptive neighborhoods of pixels, called \emph{superpixels}, that can be used for further analysis \cite{GiBo12,MuChTa16}.
Another class of methods aims to solve the segmentation problem without the knowledge of the number of present classes \cite{EsKrSa96,BaNaPi22,CaCh15}.
A third class of unsupervised hyperspectral segmentation techniques, to which also our framework belongs, needs the number of classes as prior information.
These include, besides the known general clustering methods like, e.g., $k$-means and the Gaussian mixture model that can be used for clustering of the spectra, approaches tailored for \ac{hsi} analysis.
Several approaches for spectral clustering were proposed, e.g., on a constructed anchor graph \cite{WaNiYu17,ZhYuWa19}, on a bipartite graph representation with data affinity measured by the radial basis function kernel \cite{HaKaKa18}, or based on a spectral-spatial sparse subspace clustering \cite{ZhZhZh16}.
The work by Hassanzadeh et al. \cite{HaKaKa17} introduces a multi-manifold learning framework based on contractive autoencoders.
Zhu et al. \cite{ZhChTi17} propose to combine a data fidelity term that is similar to the one in the \ac{ms} segmentation model, as described below in \cref{subsec:ms-functional}, with their developed nonlocal variant of the total variation semi-norm.
An approach that is based on density estimation and geometry learning using diffusion distance was presented by Murphy and Maggioni \cite{MuMa19}.
Zhang et al. \cite{ZhZhDu19} approach the problem by jointly reducing the dimensionality and performing the clustering using their developed unified low-rank matrix factorization.
The idea is to decompose the data matrix into a product of a projection matrix and a matrix containing lower dimensional representations of the data, while the representation matrix is further factorized into two matrices, one indicating the clusters and the other describing latent features.
More recently, Nalepa et al. \cite{NaMyIm20} proposed a 3D convolutional autoencoder that simultaneously learns a latent representation and a clustering in the latent space.
Obeid et al. \cite{ObElWe22} introduced an accelerated, balanced variant of the deep embedded clustering model \cite{XiGiFa16}.

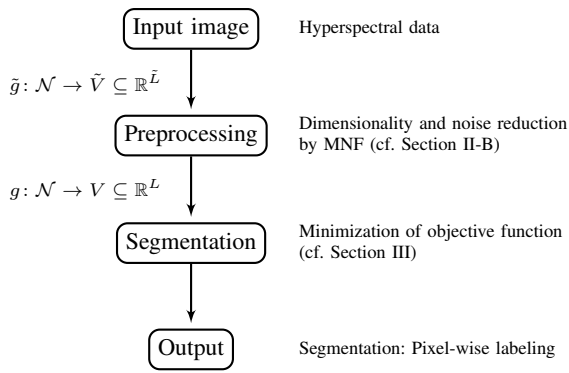
\begin{figure}
	\begin{center}
		\begin{tikzpicture}
  [font=\small,
  block/.style ={rectangle, draw=black, thick,
  align=center, rounded corners,
  minimum height=1.5em},
  line/.style ={draw, thick, -latex',shorten >=2pt},
  comment/.style ={text width=11em,align=left,
  minimum height=2em, font=\relscale{0.7}\linespread{1}\selectfont},
  io/.style ={text width=6.5em,align=left,
  minimum height=2em, font=\scriptsize\linespread{1}\selectfont}]
  \matrix [column sep=3mm,row sep=7mm]
  {
  \node [block] (input) {Input image}; &
  \node [comment] {Hyperspectral data}; \\
  \node [block] (preprocessing) {Preprocessing}; &
  \node [comment] {Dimensionality and noise reduction by MNF (cf. \cref{subsec:preprocessing})}; \\
  \node [block] (segmentation) {Segmentation}; &
  \node [comment] {Minimization of objective function (cf. \cref{sec:optimization})}; \\
  \node [block] (output) {Output}; &
  \node [comment] {Segmentation: Pixel-wise labeling}; \\
  };
  \begin{scope}[every path/.style=line]
    \path (input) -- (preprocessing) node [io, midway, left] {$\tilde{g} \colon \DiscDom \to \tilde{V} \subseteq \R^{\tilde{L}}$};
    \path (preprocessing) -- (segmentation) node [io, midway, left] {$g \colon \DiscDom \to V \subseteq \R^L$};
    \path (segmentation) -- (output);
  \end{scope}
\end{tikzpicture}%
	\end{center}
	\caption{Flowchart of our segmentation framework.}
	\label{fig:flowchart}
\end{figure}

Facing the aforementioned problems of supervised \ac{hsi} classification and due to the advantages of unsupervised \ac{hsi} segmentation, we present a novel unsupervised approach to \ac{hsi} segmentation.
Our method builds on the \ac{ms} segmentation functional.
We developed a new robust indicator function tailored for hyperspectral data that is able to handle spectral variability.
Essentially, the method is a clustering method equipped with an anisotropic distribution-dependent distance function and a spectral cluster volume regularizer that is inspired by the normalization constant of the Gaussian distribution.
Furthermore, with a second regularizer, we penalize large perimeters of segments.
The motivating assumption for this is that neighboring pixels have a high probability to belong to the same segment.
This introduces spatial information into the model only implicitly by considering the spatial gradient of the labeling function.
Part of our framework is the application of the \ac{mnf} transform to reduce the noise and dimensionality of the data.
Only there we explicitly use spatial information when estimating the noise content by considering the differences of the pixels to their respective lower right neighbors.
Hence, our framework is not a spectral approach; however, we want to point out that different from usual spectral-spatial approaches that process patches of pixels, our approach sees only a pixel plus one neighbor at a time.
\cref{fig:flowchart} shows a flowchart of the proposed framework.
It starts with a hyperspectral image as input.
The data is first preprocessed by the \ac{mnf} to reduce the dimensionality and noise.
Afterwards, the objective functional is optimized using the preprocessed data as input data.
The found minimizer yields a pixel-wise labeling of the input image as the resulting segmentation.

It should be noted that the anisotropic distance function of our indicator function is an instance of the Mahalanobis distance \cite{Ma36}.
The Mahalanobis distance has already been used in hyperspectral image processing, e.g., in Mahalanobis kernels for supervised classification \cite{LiSuLi18} and spectral dimensionality reduction \cite{LiQi20}, for detection of citrus damage \cite{GoBlMo07}, anomaly detection \cite{ZhDuZh16} or target detection \cite{Im18}.
To the best of our knowledge, we are the first who use the non-squared Mahalanobis distance for unsupervised \ac{hsi} segmentation while estimating the segments' means and covariance matrices only from the data.

Our main contributions are the following:
\begin{itemize}
	\item a novel framework for unsupervised \ac{hsi} segmentation based on the \ac{ms} segmentation functional that is equipped with a new robust indicator function;
	\item a regularization of the covariance matrix used in the Mahalanobis distance to ensure the existence and feasibility of all involved terms;
	\item an efficient fixed point iteration scheme to optimize the objective function with respect to the estimates of the means and covariances of the segments.
\end{itemize}
The remaining article is structured as follows:
\cref{sec:methodology} provides detailed information about our hyperspectral segmentation framework, i.e., the objective function of our method.
\cref{sec:optimization} explains the optimization of our functional with an alternating optimization approach.
\cref{sec:numerical-results} shows our numerical results and \cref{sec:conclusion} gives a conclusion of the article.
The source code of our segmentation framework is available at \url{https://github.com/berkels/msiplib}.

\section{Methodology}\label{sec:methodology}
\noindent
In the following, we give detailed information about our segmentation framework, starting with an introduction of the basic \ac{ms} segmentation functional in \cref{subsec:ms-functional}, followed by an explanation of the preprocessing step in \cref{subsec:preprocessing} and then presenting our modification in \cref{subsec:indicator-function}.

\subsection{Mumford-Shah segmentation functional}\label{subsec:ms-functional}
\noindent
We now describe the continuous multi-phase \ac{ms} segmentation functional \cite{MuSh89} as the basis of our model.
Note that for numerical optimization, it is later relaxed (cf. \cref{eq:zach-functional}) and discretized (cf. \cref{eq:discrete-objective-function}).
Let $\ContDom \subseteq \R^2$ be a domain, $k\in \N$ the number of segments, $\tilde{g} \colon \ContDom \to \tilde{V} \subseteq \R^{\tilde{L}}$ an input image with $\tilde{L} \in \N$ channels.
The \ac{ms} segmentation functional, denoted by $E_{\text{MS}}$, is defined as
\begin{align}\label{eq:ms-functional}
    E_{\text{MS}}[\OO_1, \dots, \OO_k] = \sum_{l=1}^k \int_{\OO_l} f_l(x) \dx + \lambda \Per(\OO_l),
\end{align}
for a partition $(\OO_1, \dots, \OO_k)$ of the image domain $\ContDom$. The subsets $\OO_1,\dots,\OO_k$ describe the segments in the image. Our goal is to obtain a segmentation of the image by finding minimizers of the \ac{ms} functional $E_{\text{MS}}$.

The functional consists of two parts: a data fidelity term, which is the integrals over the subsets $\OO_l$ of $\ContDom$, and a regularization term which is the sum of $\Per(\OO_l)$ for $l\in \{1,\dots, k\}$, the second summand in \cref{eq:ms-functional}.
The essential part of the data term is the so-called \emph{indicator functions} $f_l \colon \ContDom \to \R$ that measure how well $x \in \ContDom$ fits into the respective segment $\OO_l$.
They are required to be bounded from below.
The indicator functions $f_l$ define the notion of homogeneity for the segments and are therefore playing a crucial role in the \ac{ms} functional.
By integrating $f_l$ over $\OO_l$ for $l \in \{1,\dots, k\}$ and summing the integrals over all segments $\OO_1, \dots, \OO_k$, we obtain a measure of the goodness of a partition with respect to the chosen indicator functions.
The regularizing term $\sum_{l=1}^k \Per(\OO_l)$ deals with the perimeters of the segments $\OO_1, \dots, \OO_k$ and penalizes these if they become too large.
This term incorporates the spatial relations that are given by the image domain $\ContDom$.
The underlying assumption is that neighboring positions in the image domain have a high probability to belong to the same segment.
In particular, isolated points of a segment contribute to its perimeter and shall be avoided by this regularization, resulting in segmentations showing homogeneous regions.
The hyperparameter $\lambda > 0$ balances the data term and the regularization term in the functional.

In our case, the data term is taking care of the spectral information during segmentation.
Since we assume that each material reflects a specific spectral signature (cf. \cref{sec:introduction}), we rely only on the spectral information when clustering the spectra and performing the segmentation.
Hence, the indicator functions define the notion of spectral homogeneity in our case and should therefore respect the special characteristics of hyperspectral data.
Our proposed indicator function is introduced in \cref{subsec:indicator-function}.

To approximate minimizers of the \ac{ms} functional $E_{\text{MS}}$ (cf. \cref{eq:ms-functional}), we use Zach's convexification \cite{ZaGaFr08} of the \ac{ms} functional.
It is
\begin{align}\label{eq:zach-functional}
	J_{\zach}[u] = \sum_{l=1}^k \int_{\ContDom} u_l (x) f_l(x) \dx + \lambda \vert u_l \vert_{\text{BV}},
\end{align}
where $u = (u_1, u_2, \dots, u_k) \colon \ContDom \to \K^k$ is the so-called \emph{labeling function} that maps a point $x\in \ContDom$ to the unit simplex in $\R^k$ given as $\K^k = \{(s_1, s_2, \dots, s_k) \in \R^k \mid s_l \geq 0\; \forall \, l\in \{1,\dots,k\},\, \sum_{l=1}^k s_l = 1\}$.
To derive \cref{eq:zach-functional} from \cref{eq:ms-functional}, one introduces characteristic functions $\chi_{\OO_l}(x)$ for $\OO_l$ in the integrals in \cref{eq:ms-functional}.
Then, the integrals can be extended to integrate over the whole image domain $\ContDom$.
Relaxing the characteristic functions by replacing them by functions $u_l \colon \ContDom \to [0,1]$ such that $\sum_{l=1}^k u_l(x) = 1$ for every $x\in \ContDom$ yields \cref{eq:zach-functional}.
The last term in \cref{eq:zach-functional} is a consequence of the relation $\Per(\OO_l) = \tv{\chi_{\OO_l}}$, where $\tv{\cdot}$ is the \ac{bv} semi-norm, also known as \ac{tv}.

\subsection{Preprocessing: Dimensionality and noise reduction}\label{subsec:preprocessing}
\noindent
We use the \ac{mnf} transform \cite{GrBeSw88} as a preprocessing step to reduce the noise as well as the dimensionality of the image $\tilde{g}$.
From now on, we consider the discrete setting, i.e., for a regular grid $\DiscDom = \{1, 2, \dots, H\} \times \{1, 2, \dots, W\} \subseteq \R^2$, we consider an image $\tilde{g} \colon \DiscDom \to \tilde{V} \subseteq \R^{\tilde{L}}$ as well as a labeling function $u = (u_1, u_2, \dots, u_k) \colon \DiscDom \to \K^k$.

The \ac{mnf} is related to the often used \ac{pca} \cite{Pe01}.
But instead of returning components ordered by their explained variance as the \ac{pca}, it returns components ordered by their \ac{snr}.
To find the components that maximize the \ac{snr}, the \ac{mnf} transform assumes additive noise, i.e., that the data fulfills
\begin{align*}
	\tilde{g} = s + n,
\end{align*}
where $s \colon \DiscDom \to \R^{\tilde{L}}$ is the true signal and $n \colon \DiscDom \to \R^{\tilde{L}}$ is the noise component.
It is further supposed that the signal and noise are uncorrelated.
Consequently, we obtain
\begin{align*}
	\Sigma_D = \Sigma_S + \Sigma_N,
\end{align*}
with $\Sigma_D \in \R^{\tilde{L}\times \tilde{L}}$ being the covariance matrix of the data and $\Sigma_S\in \R^{\tilde{L}\times \tilde{L}}$ and $\Sigma_N \in \R^{\tilde{L}\times \tilde{L}}$ the covariance matrices of the signal and the noise, respectively.
The \ac{mnf} transform simultaneously diagonalizes $\Sigma_S$ and $\Sigma_N$
\begin{align*}
	W^T \Sigma_S W = \Lambda, \qquad W^T \Sigma_N W = I,
\end{align*}
to obtain a basis of generalized eigenvectors $W \in \R^{\tilde{L}\times \tilde{L}}$, which are the sought components, and the generalized eigenvalues given on the diagonal of the diagonal matrix $\Lambda \in \R^{\tilde{L}\times \tilde{L}}$, which are the \acp{snr} corresponding to the components \cite{BePhTr12}.
The noise covariance matrix is transformed to the identity matrix as this decorrelates the noise between the bands and normalizes the noise in each band to $1$ \cite{LeWoBe90}.
Hence, the $i$-th entry on the diagonal of $\Lambda$ corresponds to the $\ac{snr}$ of the $i$-th component (eigenvector) for $i\in \{1,\dots,\tilde{L}\}$.
By keeping only $L \in \N$ with $L \leq \tilde{L}$ components that show a high \ac{snr}, we reduce the dimensionality and the noise in the data at the same time. We denote the resulting image as 
\begin{align*}
	g \colon \DiscDom \to V \subseteq \R^L.
\end{align*}

Different schemes to estimate the noise are possible.
We use the difference of each pixel to its lower right neighbor based on the assumption that the true spectra are spatially constant.
Of course, one could also consider the difference to a neighbor in another direction.
All possible directions should give similar results.
The obtained differences give a band-wise noise estimate for each pixel (except for pixels at the image boundaries) from which we can compute the global noise covariance matrix.
In particular, by considering the neighboring relation of two pixels defined by $\DiscDom$, we take into account spatial relations of pixels here, which already helps to disentangle the spectral clusters slightly.
A disadvantage of this noise estimation is that it is not informed about possible segment boundaries and hence considers the differences of the spectra at segment boundaries as noise and flattens these differences slightly, what results in a few spectra that lie between the spectral clusters.
On the other hand, as described in \cite{LeWoBe90}, the \ac{mnf} is equivalent to a two stage transformation that applies a data whitening step that transforms the noise covariance matrix to the identity matrix, followed by an application of the \ac{pca}.
We inspected the data before and after the application of the \ac{mnf} by plotting the first three principal components given by the \ac{pca} of the data.
Our subjective perception is that the data whitening of the \ac{mnf} makes the data more similar to a Gaussian distribution.
To quantify our visual perception, we computed the Henze-Zirkler test statistic \cite{HeZi90} for all four datasets we test our framework on (cf. \cref{sec:numerical-results}).
The test is used to analyze if data obeys a multivariate Gaussian distribution.
It essentially computes the distance between the theoretical characteristic function of the multivariate Gaussian distribution and the empirical characteristic function of the data.
Hence, the smaller the value of the test statistic, the closer is the data distribution to the Gaussian distribution.
To make the returned values of the test statistic for raw and transformed data comparable, we applied the \ac{pca} to the raw data before computing the test statistic to reduce it to the same dimension as the data after transformation by the \ac{mnf}.
The test confirms our subjective perception and returns for the majority of the datasets' segments a better value after application of the \ac{mnf}.
As we will see later in \cref{subsec:indicator-function}, clusters that are more similar to Gaussian distributions better suit our indicator function.

In cases where the size of the \ac{hsi} data is very large, resulting in long runtimes of the \ac{mnf} transform, more efficient variants of the transform might be considered to speed up computations.
Three possible approaches are presented, for instance, in \cite{GuMiKa19}.
In cases where one deals with massive data, preventing the machine from processing the data at once and performing the \ac{mnf} transform, one can use streaming models to process the data in chunks. One such approach is presented in \cite{GuBa19}.

\subsection{Robust anisotropic indicator function}\label{subsec:indicator-function}
\noindent
As already mentioned in \cref{subsec:ms-functional}, one of the essential parts of the functional in \cref{eq:ms-functional} is the used indicator functions $f_l$ that define the notion of spectral homogeneity and therefore the distance in spectral space.
In the following, we describe our indicator function, designed to appropriately handle the characteristics of hyperspectral data.

The main challenge of hyperspectral data is, as mentioned in \cref{sec:introduction} and depicted in \cref{fig:feature-variation}, the spectral variability.
Since this means that the variation of the data in some directions in spectral space is stronger than in others, the Euclidean norm as the classic choice for the indicator function is not sophisticated enough as it is isotropic, meaning that it assumes the same variation in each direction.
Our indicator function generalizes the Euclidean norm by respecting the different strengths of variation for the respective directions and hence making it anisotropic and variability-dependent.
To this end, we consider first and second order statistics of the current segment estimates, which essentially means that the feature distributions of the segments are approximated by Gaussian distributions.

Let $g^{i,j} := g(i,j) \in V \subseteq \R^L$ for $(i,j) \in \DiscDom$.
We define our indicator function $f_l \colon \DiscDom \to \R$ for segment $l\in \{1,\dots, k\}$ as
\begin{align}\label{eq:indicator-function-bilinear}
	f_l(i,j;\mu_l, \Sigma_l) := \sqrt{(g^{i,j} - \mu_l)^T \Sigma_l^{-1} (g^{i,j} - \mu_l)} + \log \det \Sigma_l,
\end{align}
where $\mu_l \in \R^L$ and $\Sigma_l \in \R^{L \times L}$ describe the current estimates of the mean spectrum and the covariance matrix of segment $l$, respectively.
Note that the covariance matrix is symmetric by definition.
Later in this section, we will describe how we regularize the covariance matrix to ensure that the used matrix in \cref{eq:indicator-function-bilinear} is invertible.
Our indicator function consists of two parts where the first summand computes the distribution-dependent distance of a spectrum $g^{i,j}$ to the mean spectrum of the segment and the second summand is a volume regularizer of the spectral cluster.

The key idea of the first term in \cref{eq:indicator-function-bilinear} is to give a pixel a lower indicator value if it deviates from the mean spectrum of the segment in a direction where the feature distribution of the segment varies strongly and penalize small deviations in directions where the feature distribution has only a very small standard deviation.
By computing the eigendecomposition of the covariance matrix $\Sigma_l = U_l D_l^2 U_l^T$, where $U_l\in \R^{L\times L}$ is the orthogonal matrix containing the eigenvectors, which describe the principal directions of variation, and $D_l\in \R^{L\times L}$ is the diagonal matrix having the standard deviations $\sigma_{l,r} > 0$, $r\in \{1,\dots,L\}$, in the corresponding directions of variation on its diagonal, we can write \cref{eq:indicator-function-bilinear} as
\begin{align}\label{eq:indicator-function-norm}
	f_l(i,j;\mu_l, \Sigma_l) = \Vert D_l^{-1} U_l^T (g^{i,j} - \mu_l)\Vert_2 + \log \prod_{r=1}^L \sigma_{l,r}^2.
\end{align}
\cref{eq:indicator-function-norm} allows for a geometric interpretation of the indicator function:
we still measure the difference between the spectrum of the pixel at $(i,j) \in \DiscDom$ and the mean spectrum of the segment $\mu_l$; however, this difference is now projected onto the eigenvectors of $\Sigma_l$ by the multiplication with $U_l^T$ and scaled afterwards by the multiplication with the inverse of $D_l$.
These two steps mean that the vector $g^{i,j} - \mu_l$ is scaled by $1 / \sigma_{l,r}$ in the respective direction of the deviation.
Hence, if the feature distribution of the segment deviates strongly in a certain direction, the standard deviation is large, implying that its reciprocal is small and resulting in a lower indicator value for that direction.
If the feature distribution shows a very small deviation in a direction, the standard deviation is small, implying that its reciprocal is large what ends up in a large indicator value for that direction.

We propose further to use the square root in the indicator function in \cref{eq:indicator-function-bilinear} to make the estimates of $\mu_l$ and $\Sigma_l$ more robust against outliers.
As already mentioned, hyperspectral data suffers from noise.
Hence, due to the spectral variability and the noise, some spectra lie in the regions between clusters.
Such spectra have a high potential to be misclassified with the result that they would negatively influence the estimates of $\mu_l$ and $\Sigma_l$.
Moreover, the application of the \ac{mnf} transform contributes to this effect, too, caused by the current simple approach to globally estimate the noise content based on differences of neighboring pixels (cf. \cref{subsec:preprocessing}).
The problem is that pixels at segment boundaries contribute to the noise estimation for the \ac{mnf} although their difference in the spectra stems from different constituents in the scene and is therefore natural, wanted and needed for the distinction of these constituents.
In particular, the estimate of $\mu_l$, for instance, may move away from the dense region of the feature distribution and towards the outliers.
This effect is much weaker if we use the square root in \cref{eq:indicator-function-bilinear} compared to the indicator function without square root.
A downside of this idea is that, to the best of our knowledge, there exists no closed form solution for choices of $\mu_l$ and $\Sigma_l$ that let the gradient vanish in order to optimize over these quantities.
In \cref{subsec:optimization-mu-sigma}, we describe a simple fixed point iteration scheme to optimize over the two parameters.

The second term in \cref{eq:indicator-function-bilinear} is a spectral volume regularizer for each segment that tries to keep the cluster volumes small by minimizing the standard deviations.
This helps to balance the regularizing perimeter and the data term in \cref{eq:ms-functional}.
By considering the spatial neighborhood relations, the perimeter makes it possible to put spatially close pixels into the same segment that are spectrally apart.
However, doing so increases the cluster volumes and hence the standard deviations of the clusters, which makes deviations of spectra from the mean spectrum cheaper and can result in misclassifications.
From an optimization perspective, it is cheapest to let the standard deviations tend to $\infty$ if the spectral volume regularizer was omitted.
This would make the scaling factors of our indicator function tend to $0$ and hence the data term could be made arbitrarily small with $0$ as lower bound.
By keeping the standard deviations of the clusters small, this effect is avoided.
Please note that for fixed $\mu_l$ and $\Sigma_l$ the volume regularizer is only a constant shift of the indicator function.

The similarity of our model to the Gaussian mixture model is clearly visible.
Indeed, both models use first and second order statistics to describe the spectral clusters and therefore we approximate the unknown distributions with Gaussian distributions.
However, there are two substantial differences.
Our model assumes that all clusters are equally likely.
Moreover, since our model does not work with probability distributions, we have more flexibility regarding its design.
It is therefore feasible to have the square root in the indicator function.
Having this in a Gaussian distribution would make the integral over the full probability space diverge.

To deal with the occurrence of singular matrices $D_l$ and $\Sigma_l$, we propose to regularize these matrices and use the resulting matrices in the framework.
Concretely, we introduce a small $\epsilon > 0$ and replace $\sigma_{l,r}$ in $D_l$ by $\epsilon$ if $\sigma_{l,r} \leq \epsilon$.
We denote the arising regularized matrices with $\DiagReg{l}$ and $\CovReg{l} = U_l \DiagReg{l}^2 U_l^T$ and define $\Sigma_{\epsilon} := \left( \CovReg{1}, \dots, \CovReg{k} \right) \in \left( \R^{L \times L} \right)^{k}$.
This regularization step ensures that all diagonal elements of $\DiagReg{l}$ and therefore all eigenvalues of $\CovReg{l}$ are strictly positive, which makes both matrices always invertible.
Additionally, the condition of $\CovReg{l}$ is always smaller than or equal to the one of $\Sigma_l$.
With the regularization, the volume regularizer for the segments becomes
\begin{align}\label{eq:regularized-volume}
	\log \det \CovReg{l} = \log \prod_{r=1}^L \max\left\{\sigma_{l,r}^2, \epsilon^2\right\}.
\end{align}
Since we aim to minimize the objective function, the algorithm tries to make $\det \CovReg{l}$ small by reducing its eigenvalues and hence the standard deviations of the feature distribution.
However, once we have $\sigma_{l,r} \leq \epsilon$ for all $r\in \{1,\dots, L\}$, we cannot improve the objective function values further by adjusting these variables because in this case we obtain with \cref{eq:regularized-volume}
\begin{align*}
	\log \det \CovReg{l} = \log \epsilon^{2L}.
\end{align*}
Consequently, $\epsilon$ sets the minimum volume the spectral cluster of a segment can have.

In \cite{FaViCh10}, Fauvel et al. proposed to use a different regularization of the class covariance matrix for Mahalanobis kernel-based supervised \ac{hsi} classification.
The idea is to keep only the components corresponding to variances that are significantly larger than 0 and regularize these variances by adding a small constant that is greater 0.
The covariance matrix itself is estimated using labeled training data.

\section{Optimization}\label{sec:optimization}
\noindent
In this section, we describe how the objective function, i.e., the relaxed and discretized \ac{ms} segmentation functional equipped with our indicator function, is optimized with respect to its parameters.
We use an alternating minimization approach. In the following, we first give information about the overall optimization process, before we describe the subsequent steps.
These steps are the parameter initialization (cf. \cref{subsec:initialization}), the optimization over $\mu_l$ and $\Sigma_l$ for all $l\in \{1,\dots,k\}$ (cf. \cref{subsec:optimization-mu-sigma}) and the optimization over $u$ (cf. \cref{subsec:optimization-u}).
The section ends with an explanation of the used stopping criterion in \cref{subsec:stopping-criterion} and a consideration of the computational complexity in \cref{subsec:computational-complexity}.

As mentioned in \cref{subsec:indicator-function}, we have to deal with the lack of a closed form solution for $\mu_l$ and $\Sigma_l$ and the non-differentiability of the square root at $0$.
We regularize the square root by adding a small constant $\eta > 0$ to solve the latter problem.
The resulting regularized indicator function for segment $l \in \{1,\dots,k\}$ and $(i,j) \in \DiscDom$ for our discretized objective function is defined as
\begin{align}\label{eq:discretized-indicator-function}
	\begin{split}
		f_{\eta, l}(i,j;\mu_l, \Sigma_l) :=& \sqrt{\left( g^{i,j} - \mu_l \right)^T \Sigma_l^{-1} \left( g^{i,j} - \mu_l \right) + \eta}\\
		&+ \log \det \Sigma_l,
	\end{split}
\end{align}
with $\mu_l \in \R^L$ and $\Sigma_l \in \R^{L\times L}$ being positive definite but not necessarily symmetric.
Furthermore, define $f_{\eta,l}^{i,j}(\mu_l, \Sigma_l) := f_{\eta, l}(i,j;\mu_l, \Sigma_l)$ and $u_l^{i,j}:= u_l(i,j)$ for $(i,j)\in \DiscDom$ and $l \in \{1,\dots,k\}$ and $u^{i,j}:= (u_1^{i,j}, \dots, u_k^{i,j})^T \in \K^k$.
Then, we can define the relaxed and discretized objective function (cf. \cref{eq:ms-functional,eq:zach-functional,eq:discretized-indicator-function}) as
\begin{align}\label{eq:discrete-objective-function}
	\begin{split}
		E(u, \mu, \Sigma) := \sum_{l=1}^k \sum_{i=1}^H \sum_{j=1}^W u_l^{i,j} f_{\eta, l}^{i,j}(\mu_l, \Sigma_l) + \lambda \vert u_l \vert_{\text{BV}},
	\end{split}
\end{align}
with $u=(u^{i,j})_{\substack{i=1,\dots,H,\\j=1,\dots,W}} \in \left(\K^k\right)^{H\times W}$, $\mu = (\mu_1,\dots, \mu_k) \in \left(\R^L\right)^k$ and $\Sigma = (\Sigma_1, \dots, \Sigma_k) \in \left(\R^{L\times L} \right)^k$ with $\Sigma_l$ being positive definite, but not necessarily symmetric, for all $l\in \{1,\dots,k\}$.
As mentioned, optimization is done alternatingly.
Given initial estimates for $u$, $\mu$ and $\Sigma$ as described in \cref{subsec:initialization}, we compute the indicator values of our indicator function, which implies an update of $\mu$ and $\Sigma$ using $u$ as described in \cref{subsec:optimization-mu-sigma} and enables us to update $u$ afterwards following the procedure described in \cref{subsec:optimization-u}.
We alternate between these two steps of computing the indicator values (updating $\mu$ and $\Sigma$) and updating $u$ until the stopping criterion described in \cref{subsec:stopping-criterion} is met or a maximum number of iterations is reached.
Furthermore, we perform a hard assignment in each iteration after updating $u$. That is, although we allow $u^{i,j}$ to take values within $\K^k$, we project it back onto one of the vertices of the simplex by thresholding.
Concretely, we compute
\begin{align}\label{eq:thresholding-u}
	c^{i,j} := \min\left\{\argmax_{l=1,\dots,k}\left\{u_l^{i,j}\right\}\right\} \in \{1,\dots,k\},
\end{align}
as the new segment label of pixel $(i,j) \in \DiscDom$ and set $u^{i,j}$ to the resulting one-hot encoding for that pixel:
\begin{align}\label{eq:reset-u}
	u^{i,j}:= (\delta_{c^{i,j}1}, \delta_{c^{i,j}2}, \dots, \delta_{c^{i,j}k})^T \in \{0,1\}^k \cap \K^k,
\end{align}
where $\delta_{rs}$ for $r,s \in \{1,\dots,k\}$ is the Kronecker delta.
Thus, outside of \cref{subsec:optimization-u}, we assume that $u^{i,j} \in \{0,1\}^k \cap \K^k$ for each pixel $(i,j) \in \DiscDom$, i.e., $u^{i,j}$ lies on one of the vertices of the unit simplex.
The method is summarized in \cref{algo:alternating-optimization}.
\RestyleAlgo{ruled}
\begin{algorithm}[ht]
	\caption{Optimization of objective function}
	\label{algo:alternating-optimization}
	\KwIn{Hyperspectral image $\tilde{g}\colon \DiscDom \to \tilde{V}$, $\lambda > 0$, $\epsilon > 0$, $\eta > 0$;}
	\KwResult{One-hot encoding of segmentation in $u$;}
	\textbf{Dimensionality reduction:} Apply \ac{mnf} to $\tilde{g}$. (cf. \cref{subsec:preprocessing})\;
	\textbf{Initialization:} Initialize $u \in \left(\{0,1\}^k\cap \K^k\right)^{H\times W}$, $\mu \in \left(\R^L\right)^k$, $\Sigma \in \left(\R^{L \times L}\right)^k$ (cf. \cref{subsec:initialization}). Regularize $\Sigma$ with $\epsilon$ (cf. \cref{subsec:indicator-function})\;
	\While{maximum number of iterations not reached \textbf{and} stopping criterion (cf. \cref{subsec:stopping-criterion}) not met}{Update $\mu$ and $\Sigma$ and compute indicator values with $\Sigma_\epsilon$ (cf. \cref{subsec:optimization-mu-sigma})\;
	Update $u$ (cf. \cref{subsec:optimization-u})\;
	Threshold $u$ (cf. \cref{eq:thresholding-u,eq:reset-u})\;}
\end{algorithm}

\subsection{Initialization}\label{subsec:initialization}
\noindent
After normalization of the data and application of the \ac{mnf} to reduce dimensionality and noise, we initialize $u$ as well as $\mu$ and $\Sigma$ by applying the $k$-means algorithm.
Concretely, $k$-means returns a pixel labeling such that we have a class label $c \in \{1,\dots,k\}$ for each pixel.
These class labels can be translated into one-hot encodings $(\delta_{c1}, \delta_{c2}, \dots, \delta_{ck})^T \in \{0,1\}^k \cap \K^k$ as in \cref{eq:reset-u}.
By rearranging the one-hot encodings according to the pixel positions given by $\DiscDom$, we obtain an initial guess for $u$ that lies on one of the vertices of $\left(\K^k\right)^{H\times W}$.
This labeling from $k$-means is also used to compute initial values for $\mu$ and $\Sigma$ as empirical means and empirical covariance matrices of the segments according to the formulas given in \cref{eq:sample-mean,eq:sample-covariance}.
The covariance matrices are then regularized using $\epsilon$ as described before in \cref{subsec:indicator-function}.

\subsection{Optimization of mean and covariance}\label{subsec:optimization-mu-sigma}
\noindent
As mentioned in \cref{subsec:indicator-function}, the fact that $\mu_l$ and $\Sigma_l$ are present inside a square root prevented us from deriving a closed form solution for these quantities.
In the following, we describe a simple fixed point iteration scheme to optimize over these parameters efficiently.
To this end, we define the function
\begin{align*}
	h^{i,j}(\mu_l, \Sigma_l) &:= \sqrt{\left( g^{i,j} - \mu_l \right)^T \Sigma_l^{-1} \left( g^{i,j} - \mu_l \right) + \eta},
\end{align*}
for $(i,j) \in \DiscDom$ where $\mu_l \in \R^L$ and $\Sigma_l \in \R^{L\times L}$ is positive definite.
Then, for every $l \in \{1,\dots,k\}$, we obtain for the gradient of $E$ (cf. \cref{eq:discrete-objective-function}) with respect to $\mu_l$ and $\Sigma_l$
\begin{align*}
	\partial_{\mu_l} &E(u, \mu, \Sigma) = \sum_{i=1}^H \sum_{j=1}^W \frac {u_l^{i,j}}{h^{i,j}(\mu_l, \Sigma_l)} \Sigma_l^{-1}(\mu_l - g^{i,j}),\\
	\partial_{\Sigma_l} &E(u, \mu, \Sigma) =\\
	- &\sum_{i=1}^H \sum_{j=1}^W \frac {u_l^{i,j}}{2 h^{i,j}(\mu_l, \Sigma_l)} \Sigma_l^{-1} \left( g^{i,j} - \mu_l \right) \left( g^{i,j} - \mu_l \right)^T \Sigma_l^{-1}\\
	+ &\sum_{i=1}^H \sum_{j=1}^W u_l^{i,j} \Sigma_l^{-1}.
\end{align*}
Please note that the formulations of the derivatives take into account that we evaluate $E$ only at positive definite matrices $\Sigma$ that are also symmetric.
For general positive definite matrices, the derivatives deviate from the formulations above.
To compute $\partial_{\Sigma_l} E(u, \mu, \Sigma)$, we used that $\partial_{\Sigma_l} \log \det \Sigma_l = \Sigma_l^{-1}$ and that $\partial_{\Sigma_l} b^T \Sigma_l^{-1} b = -\Sigma_l^{-1} bb^T \Sigma_l^{-1}$ for $b \in \R^L$ \cite{PePe12}.
Setting the gradients to $0$ and reformulating them yields
\begin{align*}
	\mu_l &= \frac{\sum_{i=1}^H \sum_{j=1}^W \frac{u_l^{i,j}}{h^{i,j}(\mu_l, \Sigma_l)} g^{i,j}}{\sum_{i=1}^H \sum_{j=1}^W \frac{u_l^{i,j}}{h^{i,j}(\mu_l, \Sigma_l)}},\\
	\Sigma_l &=	\frac{\sum_{i=1}^H \sum_{j=1}^W \frac{u_l^{i,j}}{2 h^{i,j}(\mu_l, \Sigma_l)}\left( g^{i,j} - \mu_l \right) \left( g^{i,j} - \mu_l \right)^T}{\sum_{i=1}^H \sum_{j=1}^W u_l^{i,j}}.
\end{align*}
Since in both cases both sides of the equation depend on $\mu_l$ and $\Sigma_l$, respectively, we use these expressions to define a fixed point iteration scheme with update rules for $m\in \N_0$ given in \cref{eq:fixed-point-iteration-rule-mean,eq:fixed-point-iteration-rule-covariance}.
\begin{figure*}[!t]
	\normalsize
	\setcounter{MYtempeqncnt}{\value{equation}}
	\setcounter{equation}{9}
	\begin{align}
	\label{eq:fixed-point-iteration-rule-mean}
		\mu_l^{(m+1)} &= \frac{\sum_{i=1}^H \sum_{j=1}^W \frac{u_l^{i,j}}{h^{i,j}\left(\mu_l^{(m)}, \CovReg{l}^{(m)}\right)} g^{i,j}}{\sum_{i=1}^H \sum_{j=1}^W \frac {u_l^{i,j}}{h^{i,j}\left(\mu_l^{(m)}, \CovReg{l}^{(m)}\right)}},\\
		\Sigma_l^{(m+1)} &= \frac{\sum_{i=1}^H \sum_{j=1}^W \frac {u_l^{i,j}}{2 h^{i,j}\left(\mu_l^{(m)}, \CovReg{l}^{(m)}\right)} \left( g^{i,j} - \mu_l^{(m+1)} \right) \left( g^{i,j} - \mu_l^{(m+1)} \right)^T}{\sum_{i=1}^H \sum_{j=1}^W u_l^{i,j}}
	\label{eq:fixed-point-iteration-rule-covariance}
	\end{align}
	\setcounter{equation}{\value{MYtempeqncnt} + 2}
	\hrulefill
	\vspace*{4pt}
\end{figure*}
Please note that on the left hand side of \cref{eq:fixed-point-iteration-rule-covariance} we have the \emph{un}regularized covariance matrix.
It has to be regularized with an $\epsilon > 0$ as introduced in \cref{subsec:indicator-function} by projecting it onto the admissible set of positive definite matrices to ensure invertibility before the next step of the fixed point iteration can be performed.

One can motivate the iteration scheme for $\mu_l$ geometrically: find a mean spectrum $\mu_l$ that lies within the dense region of the feature distribution by iteratively reducing the influence of points that are far away.
Start with an initial estimate, compute the distances of all spectra in the segment to the mean, and use the reciprocal distances as weights to compute a new weighted mean.
Repeat the procedure until convergence.
This drags $\mu_l$ inside the dense region by reducing the influence of more distant spectra.
\cref{eq:fixed-point-iteration-rule-mean} shows clearly the same structure as the geometric motivation.
In the numerator, $u_l^{i,j}$ defines the segment affiliation of each pixel and decides which pixels have to contribute to the mean of segment $l$.
The fraction has the distance of the pixel at $(i,j)$ to the current mean as its denominator, given by $h^{i,j}\left(\mu_l^{(m)}, \CovReg{l}^{(m)}\right)$.
The fraction is used to weight its spectrum $g^{i,j}$ in the new iterate for $\mu_l$.
The denominator of the outer fraction normalizes the weights to sum up to $1$.

As initial estimates for the first outer iteration of \cref{algo:alternating-optimization}, we propose to use the empirical mean and empirical covariance:
\begin{align}\label{eq:sample-mean}
	\mu_l^{(0)} &= \frac{\sum_{i=1}^H \sum_{j=1}^W u_l^{i,j} g^{i,j}}{\sum_{i=1}^H \sum_{j=1}^W u_l^{i,j} },\\
	\Sigma_l^{(0)} &= \frac{\sum_{i=1}^H \sum_{j=1}^W u_l^{i,j} \left( g^{i,j} - \mu_l^{(0)} \right) \left( g^{i,j} - \mu_l^{(0)} \right)^T}{\sum_{i=1}^H \sum_{j=1}^W u_l^{i,j} - 1}.
	\label{eq:sample-covariance}
\end{align}
Once the fixed point iteration scheme was applied for the first time, i.e., in the first outer iteration, its output can be used as an initial guess for the next application in the following outer iteration.
Please note as above that \cref{eq:sample-covariance} yields the \emph{un}regularized $\Sigma_l^{(0)}$ that needs to be regularized to obtain $\CovReg{l}^{(0)}$.

We stop the fixed point iteration once a maximum number $m_{\mu,\Sigma}^{\max} \in \N$ of iterations is reached or the expression
\begin{align*}
	\left\Vert \mu_l^{(m+1)} - \mu_l^{(m)}\right\Vert_2
	&+ \left\Vert D_{\epsilon, l}^{(m+1)} - D_{\epsilon, l}^{(m)} \right\Vert_F\\
	&+ \left\Vert U_{l}^{(m+1)} - U_{l}^{(m)} \right\Vert_F
\end{align*}
is below a certain threshold.
Here, the matrices $D_{\epsilon, l}^{(m)}$ and $U_{l}^{(m)}$ are given by the eigenvalue decomposition $\CovReg{l}^{(m)} = U_{l}^{(m)} D_{\epsilon, l}^{(m)} \left(U_{l}^{(m)}\right)^T$.

\subsection{Optimization of labeling function}\label{subsec:optimization-u}
\noindent
We use the \ac{pdhg} method proposed by Chambolle and Pock \cite{ChPo11} to optimize $E$ with respect to $u$.
It is especially suitable to solve convex optimization problems of the form
\begin{align*}\label{eq:pdhg-optimization-problem}
	\min_{u \in \R^n} F(u) + G(Au),
\end{align*}
for closed, proper and convex functions $F \colon \R^n \to \R$ and $G \colon \R^d \to \R$, as well as a linear function $A \colon \R^n \to \R^d$.
In our case, $F$ is given as the data term of our objective functional (cf. \cref{eq:discrete-objective-function}), i.e., the sum of the indicator values multiplied by the values of $u$ and summed over all segments plus the convex indicator function of the admissible set $\left(\K^k\right)^{H\times W}$ that outputs $\infty$ if $u$ is outside of $\left(\K^k\right)^{H\times W}$ and $0$ otherwise.
The linear function $A$ is the forward finite differences operator with step size $h = 1 / (\max(H, W) - 1)$ that approximates for each pixel $(i,j) \in \DiscDom$ fulfilling $i\neq H$ and $j \neq W$ the class-wise gradient $\nabla u(i,j)$ by comparing $u(i,j)$ to $u(i,j+1)$ and $u(i+1,j)$, respectively.
At boundary pixels where forward finite differences point outside of the domain, we set the derivative in the corresponding direction to $0$.
The function $G$ computes the sum over the pixel-wise Euclidean norm of the pixel-wise gradient of $u$, which equals the discrete total variation of $u$.

Let $\tau > 0$ and $\sigma > 0$.
The \ac{pdhg} algorithm is given in \cref{eq:pdhg-method}.
\RestyleAlgo{ruled}
\begin{algorithm}[ht]
	\caption{\acl{pdhg} method}
	\label{eq:pdhg-method}
	\KwIn{$\tau > 0, \sigma > 0, \LevelSetFuncd^{(0)} \in \R^n$;}
	\KwResult{$\LevelSetFuncd^{(m+1)} \in \R^n$;}
	\textbf{Initialization:} $\bar{\LevelSetFuncd}^{(0)} = \LevelSetFuncd^{(0)} \in \mathbb{R}^n, \DualVard^{(0)} \in \mathbb{R}^d$ and $\theta \in [0,1]$\;
	\While{maximum number of iterations not reached \textbf{and} $\Vert \LevelSetFuncd^{(m+1)} - \LevelSetFuncd^{(m)} \Vert_{\infty} \geq t$ for a threshold $t > 0$}{
		$\DualVard^{(m+1)} = \ProximalMapping_{\sigma G^*}\left( \DualVard^{(m)} + \sigma A \bar{\LevelSetFuncd}^{(m)} \right)$\;
		$\LevelSetFuncd^{(m+1)} = \ProximalMapping_{\tau F}\left( \LevelSetFuncd^{(m)} - \tau A^T \DualVard^{(m+1)} \right)$\;
		$\bar{\LevelSetFuncd}^{(m+1)} = \LevelSetFuncd^{(m+1)} + \theta \left( \LevelSetFuncd^{(m+1)} - \LevelSetFuncd^{(m)} \right)$\;
	}
\end{algorithm}
For $(i,j)\in \DiscDom$ define $f_{\eta}^{i,j}:= (f_{\eta, 1}^{i,j}(\mu_1, \Sigma_1), \dots, f_{\eta, k}^{i,j}(\mu_k, \Sigma_k))^T \in \R^k$.
Furthermore, let $f_{\eta} := \left(f_{\eta}^{i,j}\right)_{\substack{i=1,\dots,H,\\j=1,\dots,W}} \in \left(\R^k\right)^{H\times W}$ be the tensor containing the indicator values of all pixels.
Then, the involved proximal mappings can be derived in a closed form, i.e.,
\begin{align*}
	\ProximalMapping_{\tau F} (u) &= \Projection_{\K^k}\left(u - \frac \tau \lambda f_\eta\right),\\
	\ProximalMapping_{\sigma G^*} (p) &= \Projection_{\overline{\mathcal{B}_{2}}}(p),
\end{align*}
where $\Projection_{\K^k}$ is the pixel-wise projection onto the unit simplex $\K^k$ and $\Projection_{\overline{\mathcal{B}_{2}}}$ the pixel-wise projection onto the closed unit ball with respect to the Euclidean norm in two dimensions.
An efficient algorithm to compute $\Projection_{\K^k}$ can be found in \cite{WaCa13}.

\subsection{Stopping criterion}\label{subsec:stopping-criterion}
\noindent
We stop the outer iteration in \cref{algo:alternating-optimization} once the expression
\begin{align*}
	\sum_{l=1}^k \frac{\sum_{i=1}^{H}\sum_{j=1}^{W} (u^{(m)})_l^{i,j}}{HW} \left\Vert \mu_l^{(m)} - \mu_l^{(m-1)} \right\Vert_\infty < t,
\end{align*}
with $m \in \N$ being the current iteration, is true for a threshold $t > 0$ or when $m = m^{\text{max}}$ for a constant $m^{\text{max}}\in \N$.
The idea is to stop the iteration when the resulting segmentation, i.e., $u$ after thresholding, does not change substantially anymore or when a maximum number of iterations is reached.
We use the mean spectra of the segments to measure the change in $u$.
The weighting by the cardinality of the segments is introduced to reduce the influence of changes of labels in segments with only a few pixels as in such cases the influence of a single pixel on the mean spectrum is stronger than in segments with many pixels.
Therefore, using such a weighting, it is avoided that only a few pixels in small segments keep the algorithm iterating because their labels are alternating between two classes.

\subsection{Computational complexity}\label{subsec:computational-complexity}
\noindent
The preprocessing step by the \ac{mnf} transform has a complexity of $\OO(\tilde{L}^3 + \tilde{L}^2 HW)$ due to the eigendecomposition with $\OO(\tilde{L}^3)$ and the projection onto the found components with complexity $\OO(\tilde{L}^2HW)$ \cite{GuMiKa19}.
One application of $k$-means costs $\OO(kHWm_{k\text{-means}}^{\text{max}})$ \cite{MaRaSc08} where $m_{k\text{-means}}^{\text{max}} \in \N$ is the maximum number of iterations chosen for $k$-means.
The initialization of $\mu$ and $\Sigma_{\epsilon}$, i.e., including the regularization, has a cost of $\OO(k[L^3 + L^2HW])$ where the $L^2HW$-term stems from the computation of the covariance matrices and the $L^3$-term from the eigendecomposition for the regularization.
In fact, the eigendecompositions that are computed for the regularization of the covariance matrices allow to multiply a vector with the inverse regularized covariance matrix $\Sigma_{\epsilon, l}^{-1}$ using only $\OO(L^2)$ operations.
One update of $\mu$ and $\Sigma$ has a complexity of $\OO(m_{\mu,\Sigma}^{\text{max}} k [L^3 + L^2HW])$ as each iteration of the fixed point iteration scheme (cf. \cref{subsec:optimization-mu-sigma}) requires a regularization and therefore eigendecomposition of $\Sigma_l$ for all $l \in \{1,\dots, k\}$ with $\OO(L^3)$ steps and $HW$ outer products in the numerator of \cref{eq:fixed-point-iteration-rule-covariance} with $\OO(L^2)$ steps. As in \cref{subsec:optimization-mu-sigma}, $m_{\mu,\Sigma}^{\text{max}}$ describes the maximum number of iterations of the fixed point iteration scheme.
One update of $u$ (cf. \cref{subsec:optimization-u}) costs $\OO(kL^2HW + m_{\text{PDHG}}^{\max} k HW \log k)$ where $m_{\text{PDHG}}^{\max} \in \N$ is the maximum number of iterations performed by the \ac{pdhg} method.
The first summand stems from the computation of the indicator values $f_{\eta}$.
The second summand accounts for the actual \ac{pdhg} application and includes the projection of $u$ onto $\K^k$ with complexity $\OO(k HW \log k)$ \cite{WaCa13} and the projection of $p$ onto the unit ball with complexity $\OO(kHW)$.
The updates of $\mu$ and $\Sigma$ as well as the update of $u$ are computed at most $m^{\max}$ times with $m^{\max}$ being the maximum number of outer iterations as above.
We arrive at an overall computational complexity of $\OO(\tilde{L}^3 + \tilde{L}^2HW + m_{k\text{-means}}^{\text{max}} kHW + m^{\text{max}} [m_{\mu,\Sigma}^{\text{max}} k \{L^3 + L^2HW\} + m_{\text{PDHG}}^{\max} kHW \log k])$.

\section{Numerical results}\label{sec:numerical-results}
\noindent
Tests of our model were performed on four publicly available datasets from the remote sensing community: \ac{ip}, \ac{ksc}, \ac{pu} and \ac{sa}.

\subsection{Datasets}
\noindent
\urldef\urldata\url{http://www.ehu.eus/ccwintco/index.php?title=Hyperspectral_Remote_Sensing_Scenes}%
Brief descriptions of the datasets\footnote{Datasets available under \urldata} that were used for evaluation are given in the following:
\begin{enumerate}
	\item \emph{\acl{ip}}: this dataset, that was acquired with an \ac{aviris} sensor in north-western Indiana, has a size of $145 \times 145$ pixels and contains $220$ spectral channels per pixel. The ground truth divides the image into 16 land cover classes. We use the version where the water absorption bands were discarded, resulting in $200$ channels per pixel.
	\item \emph{\acl{ksc}}: this dataset was acquired with an \ac{aviris} sensor over the \acl{ksc} in Florida. The $512 \times 614$ pixels contain 224 channels each. The water absorption and noisy channels were discarded, resulting in 176 channels for our analysis. The ground truth describes 13 classes of land cover types.
	\item \emph{\acl{pu}}: the data was acquired with a ROSIS sensor over Pavia in northern Italy. There are $103$ clean channels for each of the $610 \times 340$ pixels. The ground truth distinguishes between 9 land cover types.
	\item \emph{\acl{sa}}: the last dataset is again an \ac{aviris} dataset, acquired over the Salinas Valley in California. It is built of $512 \times 217$ pixels with $224$ channels each. The ground truth divides the data into 16 classes of land cover types. We test on the version of the dataset where 20 water absorption bands were removed.
\end{enumerate}

\subsection{Assessment of segmentations}
\noindent
To assess our obtained segmentations, we compare the segmentation map to the corresponding ground truth of the respective dataset.
Since the labeling of the segments in the segmentation map does not necessarily correspond to the ground truth labels, we find the best matching of the labels by computing the confusion matrix and letting the Hungarian algorithm \cite{Ku55} find the permutation that leads to the highest values on the diagonal and therefore the best matching in this sense.
Based on this confusion matrix, we compute the \ac{oa}, the \ac{aa} and the \ac{kc} \cite{Co60}. \ac{oa} is defined as the fraction of correctly classified pixels, \ac{aa} describes the average over the class accuracies and \ac{kc} measures the agreement of two classifications taking into account a possible agreement by chance.

\subsection{Experimental setup}\label{subsec:experimental-setup}
\noindent
Before applying our segmentation framework, we normalize the raw data to have intensity values in each channel and for each pixel in $[0,1]$ by subtracting the minimum intensity and dividing by the resulting maximum.
Since this transformation is linear, it only changes the value range but not the structure of the data.

We compare our model with the classic clustering approaches $k$-means, \ac{gmm} and \ac{bgm} and use their scikit-learn implementations with default parameter settings, except for the number of maximum iterations for the \ac{bgm} that we increased to 1000.
Furthermore, we compare our model with the \ac{ms} model equipped with the squared Euclidean norm (MS-2) as the classic indicator function that we generalize with our approach.
Finally, a comparison with three state-of-the-art unsupervised hyperspectral segmentation methods is done, namely, the sequential spectral clustering approach combined with the RBF kernel (SSC) \cite{HaKaKa18}, a \ac{ms} based segmentation functional equipped with a nonlocal total variation regularization (NLTV) \cite{ZhChTi17} and a 3D convolutional autoencoder combined with a clustering layer \acused{3dcae}(\ac{3dcae}) that trains the clustering and the autoencoder simultaneously \cite{NaMyIm20}.
Please note that NLTV and our framework are both based on the \ac{ms} segmentation functional.
Some of the main differences are that in NLTV the labeling function $u$ is squared, that they use a nonlocal gradient of $u$ in the total variation and that their indicator function is a combination of cosine distance and Euclidean distance using only first order statistics.
All methods were given the number of segments $k$ as prior information.
As we had no access to the implementations of SSC and NLTV, we report the tabulated scores of their applications to the raw datasets from the literature.
The remaining methods were applied to the raw datasets and the datasets after preprocessing by the \ac{mnf}.
Furthermore, to evaluate the effect of the \ac{mnf} transform, we also applied an unsupervised feature learning approach based on a 3D convolutional autoencoder \cite{MeJiGe19} \acused{3dfeats}(\ac{3dfeats}) to the datasets \ac{ip}, \ac{pu} and \ac{sa} and applied all methods we had access to to the learned features.
We could not apply the feature learning to \ac{ksc} because the authors of the framework use a specific network architecture for \ac{aviris} data and add the noisy and water absorption channels to the data that were originally discarded and fill these with zeros.
However, for the \ac{ksc} dataset it is not documented which channels were discarded.
For each dataset, the feature learning yields 432 features per pixel.
The \ac{mnf} is applied in such a way that it keeps the number of features as specified in \cref{tab:model-parameters}.
Please also note that \ac{3dcae} requires at least 9 channels in order to be able to process the data as it reduces the number of channels by 8.
We therefore add channels filled with zeros to the \ac{mnf}-transformed data until the datasets have 9 channels before applying \ac{3dcae}.
The architecture and training of the network are as proposed by the authors.
In particular, the autoencoder is trained for 25 epochs.
The latent dimension is set to 25 for the raw data and the learned features and equal to the input dimension in case of \ac{mnf}-preprocessed data.
The batch sizes are chosen as stated in \cref{tab:model-parameters}.
\begin{table*}[ht]
	\centering
	\begin{tabular}{|lc|c|c|c|c|c|c|}
		\hline
		& & \multicolumn{3}{|c|}{\textbf{\ac{ip}}} & \multicolumn{3}{|c|}{\textbf{\ac{ksc}}}\\
		\hline
		 & $k$ & \multicolumn{3}{|c|}{16} & \multicolumn{3}{|c|}{13}\\
		\hline
		\textbf{\ac{mnf}} & $L$ & \multicolumn{3}{|c|}{8} & \multicolumn{3}{|c|}{6}\\
		\hline
		& & \textbf{raw} & \textbf{\ac{mnf}} & \textbf{\ac{3dfeats}} & \textbf{raw} & \textbf{\ac{mnf}} & \textbf{\ac{3dfeats}}\\
		\hline
		\textbf{MS-2} & $\lambda$ & 0.0021 & 0.33 & 0.00023 & 0.0003 & 0.094 & - \\
		\hline
		\multirow{2}{*}{\textbf{Ours}} & $\lambda$ & 0.078 & 0.24 & 0.21 & 0.0014 & 0.0005 & - \\
		& $\epsilon$ & 0.008 & 0.125 & 0.0025 & 0.18 & 0.5 & - \\
		\hline
		\textbf{\ac{3dcae}} & Batch size & \multicolumn{3}{c|}{841} & \multicolumn{3}{c|}{1024}\\
		\hline
		\multicolumn{8}{c}{\phantom{row}}\\
		\hline
		& & \multicolumn{3}{|c|}{\textbf{\ac{pu}}} & \multicolumn{3}{|c|}{\textbf{\ac{sa}}}\\
		\hline
		 & $k$ & \multicolumn{3}{|c|}{9} & \multicolumn{3}{|c|}{16}\\
		\hline
		\textbf{\ac{mnf}} & $L$ & \multicolumn{3}{|c|}{5} & \multicolumn{3}{|c|}{7}\\
		\hline
		& & \textbf{raw} & \textbf{\ac{mnf}} & \textbf{\ac{3dfeats}} & \textbf{raw} & \textbf{\ac{mnf}} & \textbf{\ac{3dfeats}}\\
		\hline
		\textbf{MS-2} & $\lambda$ & 0.0009 & 0.0565 & $6.85 \cdot 10^{-5}$ & 0.0009 & 0.135 & $3.75 \cdot 10^{-5}$\\
		\hline
		\multirow{2}{*}{\textbf{Ours}} & $\lambda$ & 0.0101 & 0.008 & 0.155 & 0.1 & 0.15 & 0.00042 \\
		& $\epsilon$ & 0.03 & 0.3 & $5.46\cdot 10^{-5}$ & 0.0156 & 0.115 & 0.112 \\
		\hline
		\textbf{\ac{3dcae}} & Batch size & \multicolumn{3}{c|}{1220} & \multicolumn{3}{c|}{992}\\
		\hline
	\end{tabular}
	\caption{Chosen model parameters. The number of segments $k$ is given by the ground truth.}
	\label{tab:model-parameters}
\end{table*}
The \ac{ms} model with the Euclidean norm as indicator function as well as our proposed indicator function is run with the same stopping criteria, thresholds and maximum numbers of iterations for the \ac{pdhg} algorithm and the outer iterations.
The chosen values are stated in \cref{tab:independent-ms-params}.
The table contains also the other image-independent hyperparameters of our framework.
The number of components $L$ that are kept by the \ac{mnf} was manually analyzed in a first step and fixed to the values given in \cref{tab:model-parameters}.
These values were used across all experiments.
\begin{table}[ht]
	\centering
	\begin{tabular}{|c|c|c|}
		\hline
		\multicolumn{3}{|c|}{$\mathbf{\eta}$}\\
		\hline
		\multicolumn{3}{|c|}{$10^{-2}$}\\
		\hline
		\multicolumn{3}{c}{\phantom{row}}\\
		\hline
		\multicolumn{3}{|c|}{\textbf{Stopping thresholds}}\\
		Outer iterations & PDHG iterations & Fixed point iterations\\
		\hline
		$10^{-6}$ & $10^{-6}$ & $10^{-5}$\\
		\hline
		\multicolumn{3}{c}{\phantom{row}}\\
		\hline
		\multicolumn{3}{|c|}{\textbf{Maximum number of iterations}}\\
		Outer iterations & PDHG iterations & Fixed point iterations\\
		\hline
		$20$ & $1000$ & $20$\\
		\hline
	\end{tabular}
	\caption{Image-independent parameters of our framework.}
	\label{tab:independent-ms-params}
\end{table}
The hyperparameters $\lambda$ of the \ac{ms} functional (cf. \cref{subsec:ms-functional}) and $\epsilon$ of our indicator function (cf. \cref{subsec:indicator-function}) that are stated in \cref{tab:model-parameters} were tuned using Optuna \cite{AkSaYa19} and a fine-tuning using grid search around the best Optuna parameters afterwards, everything with an arbitrary but fixed seed for the $k$-means initialization that was used for the squared Euclidean norm and our proposed indicator function.
Using these hyperparameters, every method was applied ten times to each dataset with ten different and randomly sampled seeds.

\subsection{Results}\label{subsec:results}%
\begin{figure*}[ht]
	\centering
	\subfloat[Ground truth]{\label{img:ip-raw-gt}\includegraphics[width=0.22\textwidth]{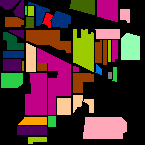}}
    \quad
	\subfloat[$k$-means (0.369)]{\label{img:ip-raw-kmeans-seg}\includegraphics[width=0.22\textwidth]{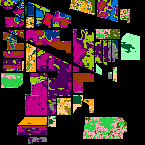}}
    \quad
	\subfloat[GMM (0.399)]{\label{img:ip-raw-gmm-seg}\includegraphics[width=0.22\textwidth]{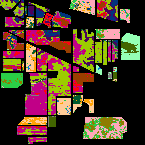}}
    \quad
	\subfloat[BGM (0.399)]{\label{img:ip-raw-bgm-seg}\includegraphics[width=0.22\textwidth]{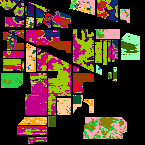}}
    \\
	\subfloat[3D-CAE (0.418)]{\label{img:ip-raw-3dcae-seg}\includegraphics[width=0.22\textwidth]{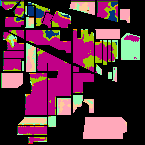}}
    \quad
	\subfloat[MS-2 (0.497)]{\label{img:ip-raw-2norm-seg}\includegraphics[width=0.22\textwidth]{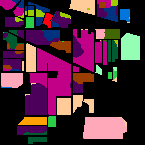}}
    \quad
	\subfloat[Ours (0.558)]{\label{img:ip-raw-an2norm-seg}\includegraphics[width=0.22\textwidth]{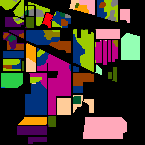}}
	\caption{Ground truth and segmentations of raw \ac{ip} data produced by the competing methods.}
	\label{fig:ip-raw-gt-seg}
\end{figure*}

\begin{figure*}[ht]
	\centering
	\subfloat[Ground truth]{\label{img:ip-mnf-gt}\includegraphics[width=0.22\textwidth]{input/ip_gt.png}}
    \quad
	\subfloat[$k$-means (0.504)]{\label{img:ip-mnf-kmeans-seg}\includegraphics[width=0.22\textwidth]{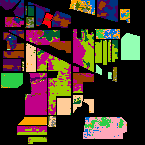}}
    \quad
	\subfloat[GMM (0.510)]{\label{img:ip-mnf-gmm-seg}\includegraphics[width=0.22\textwidth]{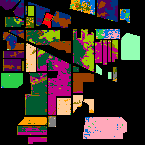}}
    \quad
	\subfloat[BGM (0.507)]{\label{img:ip-mnf-bgm-seg}\includegraphics[width=0.22\textwidth]{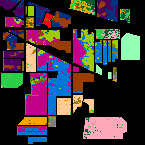}}
    \\
	\subfloat[3D-CAE (0.475)]{\label{img:ip-mnf-3dcae-seg}\includegraphics[width=0.22\textwidth]{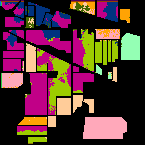}}
    \quad
	\subfloat[MS-2 (0.577)]{\label{img:ip-mnf-2norm-seg}\includegraphics[width=0.22\textwidth]{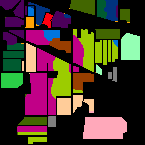}}
    \quad
	\subfloat[Ours (0.673)]{\label{img:ip-mnf-an2norm-seg}\includegraphics[width=0.22\textwidth]{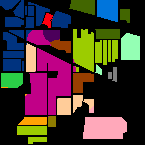}}
	\caption{Ground truth and segmentations of \ac{mnf}-transformed \ac{ip} data produced by the competing methods.}
	\label{fig:ip-mnf-gt-seg}
\end{figure*}

\begin{figure*}[ht]
	\centering
	\subfloat[Ground truth]{\label{img:ip-3dcae-fl-gt}\includegraphics[width=0.22\textwidth]{input/ip_gt.png}}
    \quad
	\subfloat[$k$-means (0.401)]{\label{img:ip-3dcae-fl-kmeans-seg}\includegraphics[width=0.22\textwidth]{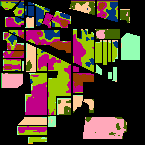}}
    \quad
	\subfloat[GMM (0.469)]{\label{img:ip-3dcae-fl-gmm-seg}\includegraphics[width=0.22\textwidth]{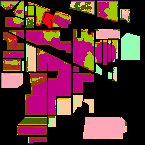}}
    \quad
	\subfloat[BGM (0.410)]{\label{img:ip-3dcae-fl-bgm-seg}\includegraphics[width=0.22\textwidth]{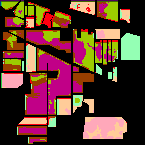}}
    \\
	\subfloat[3D-CAE (0.415)]{\label{img:ip-3dcae-fl-3dcae-seg}\includegraphics[width=0.22\textwidth]{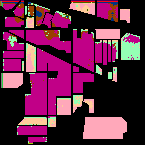}}
    \quad
	\subfloat[MS-2 (0.481)]{\label{img:ip-3dcae-fl-2norm-seg}\includegraphics[width=0.22\textwidth]{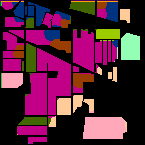}}
    \quad
	\subfloat[Ours (0.497)]{\label{img:ip-3dcae-fl-an2norm-seg}\includegraphics[width=0.22\textwidth]{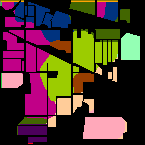}}
	\caption{Ground truth and segmentations of the learned \ac{3dfeats} features of the \ac{ip} data produced by the competing methods.}
	\label{fig:ip-3dcae-fl-gt-seg}
\end{figure*}

\begin{figure*}[ht]
	\centering
	\subfloat[Ground truth]{\label{img:ksc-raw-gt}\includegraphics[width=0.22\textwidth]{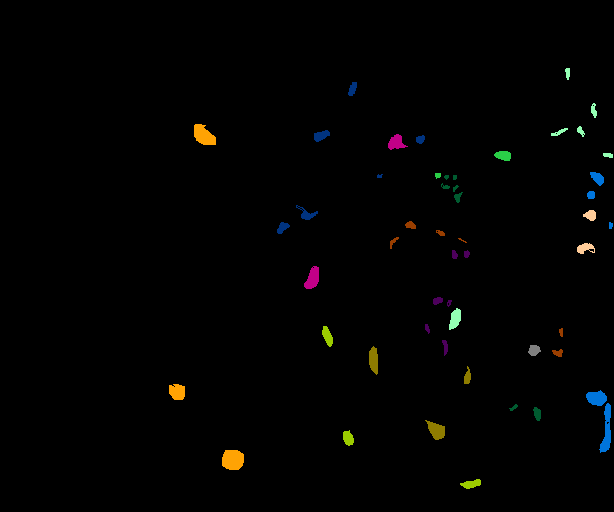}}
    \quad
	\subfloat[$k$-means (0.178)]{\label{img:ksc-raw-kmeans-seg}\includegraphics[width=0.22\textwidth]{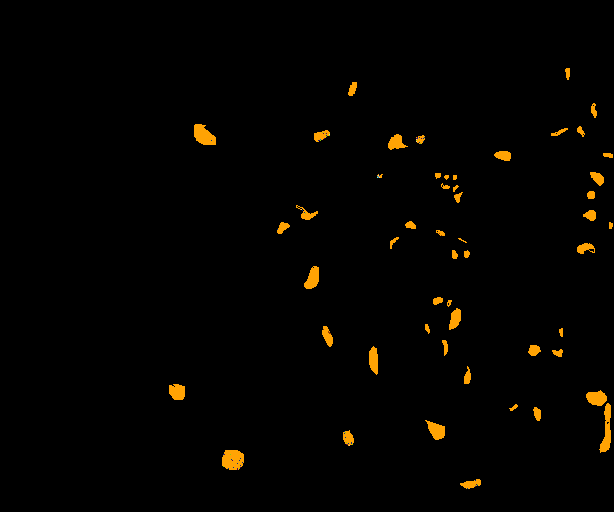}}
    \quad
	\subfloat[GMM (0.178)]{\label{img:ksc-raw-gmm-seg}\includegraphics[width=0.22\textwidth]{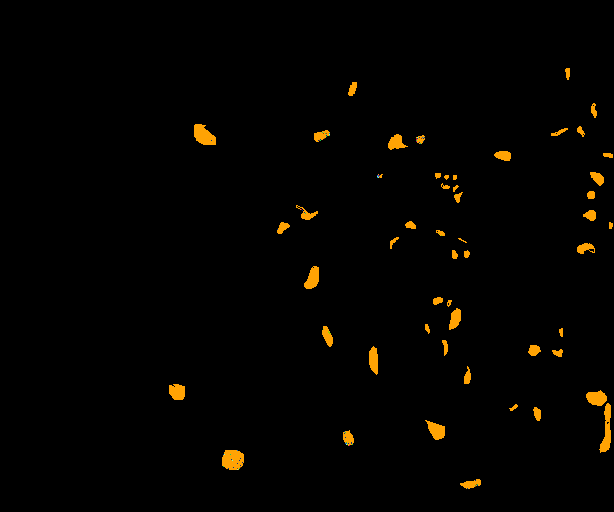}}
    \quad
	\subfloat[BGM (0.178)]{\label{img:ksc-raw-bgm-seg}\includegraphics[width=0.22\textwidth]{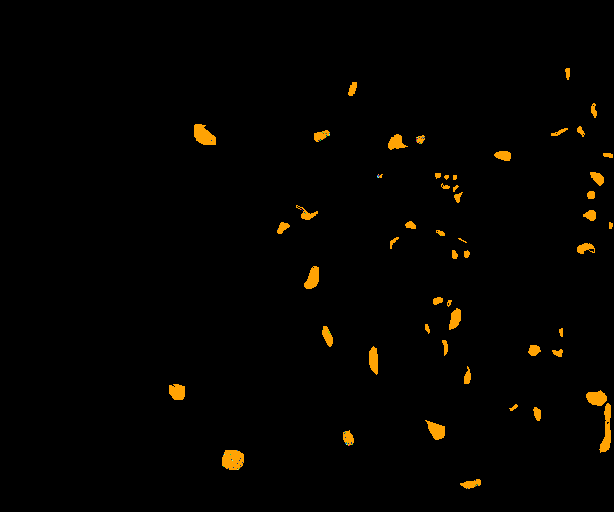}}
    \\
	\subfloat[3D-CAE (0.191)]{\label{img:ksc-raw-3dcae-seg}\includegraphics[width=0.22\textwidth]{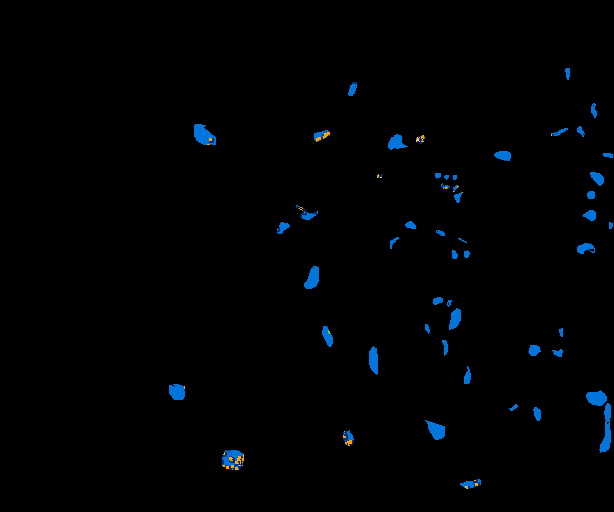}}
    \quad
	\subfloat[MS-2 (0.179)]{\label{img:ksc-raw-2norm-seg}\includegraphics[width=0.22\textwidth]{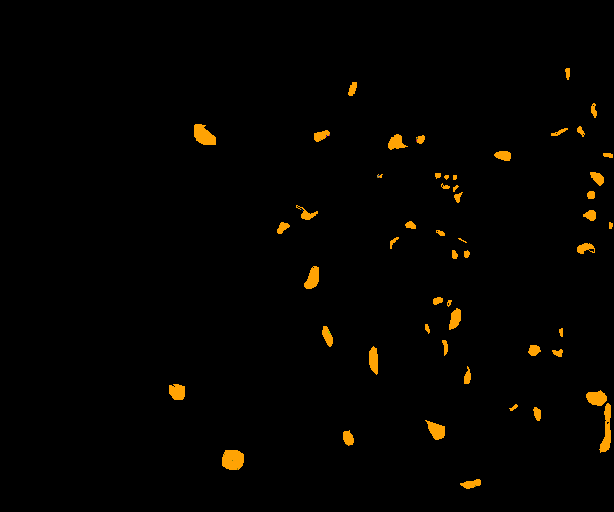}}
    \quad
	\subfloat[Ours (0.180)]{\label{img:ksc-raw-an2norm-seg}\includegraphics[width=0.22\textwidth]{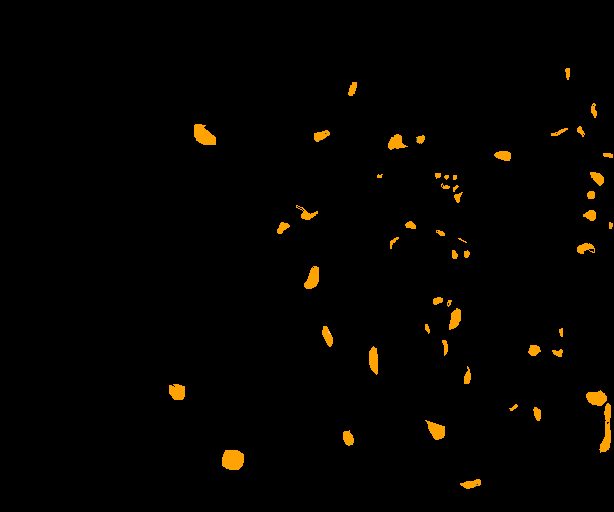}}
	\caption{Ground truth and segmentations of raw \ac{ksc} data produced by the competing methods.}
	\label{fig:ksc-raw-gt-seg}
\end{figure*}

\begin{figure*}[ht]
	\centering
	\subfloat[Ground truth]{\label{img:ksc-mnf-gt}\includegraphics[width=0.22\textwidth]{input/ksc_gt.png}}
    \quad
	\subfloat[$k$-means (0.623)]{\label{img:ksc-mnf-kmeans-seg}\includegraphics[width=0.22\textwidth]{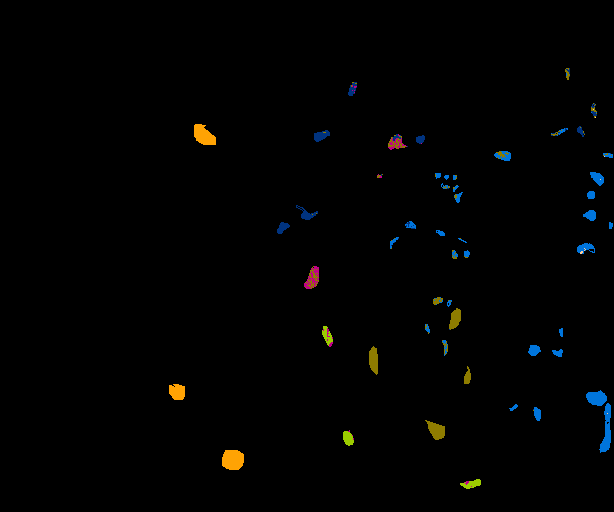}}
    \quad
	\subfloat[GMM (0.605)]{\label{img:ksc-mnf-gmm-seg}\includegraphics[width=0.22\textwidth]{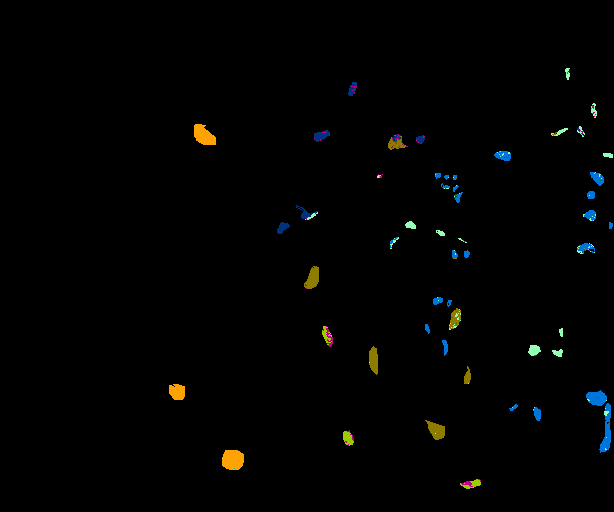}}
    \quad
	\subfloat[BGM (0.613)]{\label{img:ksc-mnf-bgm-seg}\includegraphics[width=0.22\textwidth]{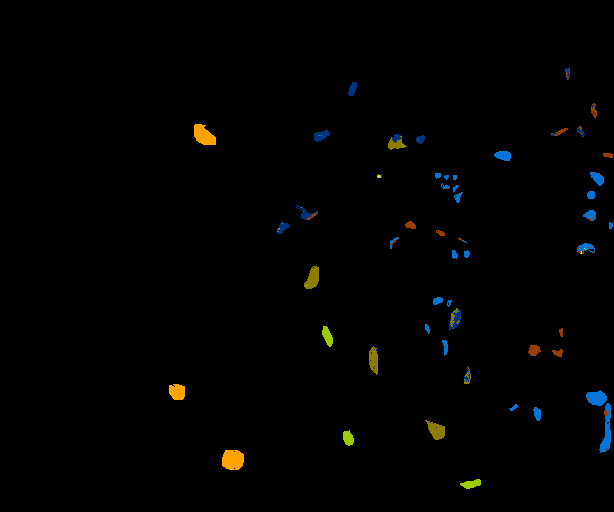}}
    \\
	\subfloat[3D-CAE (0.435)]{\label{img:ksc-mnf-3dcae-seg}\includegraphics[width=0.22\textwidth]{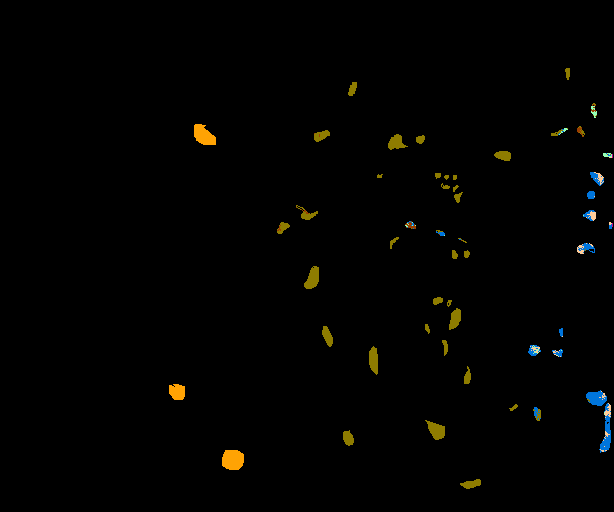}}
    \quad
	\subfloat[MS-2 (0.676)]{\label{img:ksc-mnf-2norm-seg}\includegraphics[width=0.22\textwidth]{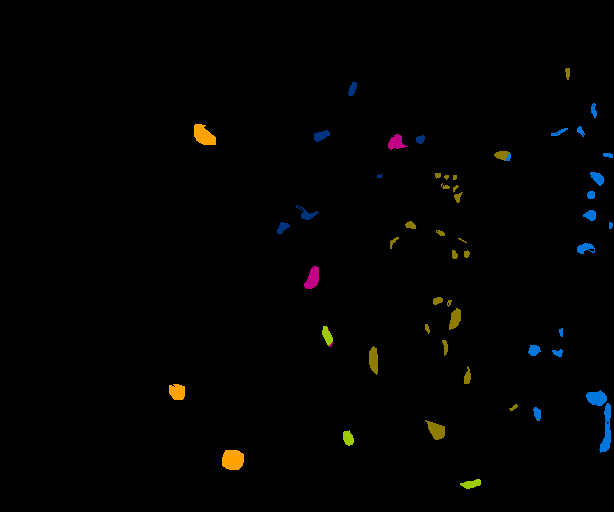}}
    \quad
	\subfloat[Ours (0.669)]{\label{img:ksc-mnf-an2norm-seg}\includegraphics[width=0.22\textwidth]{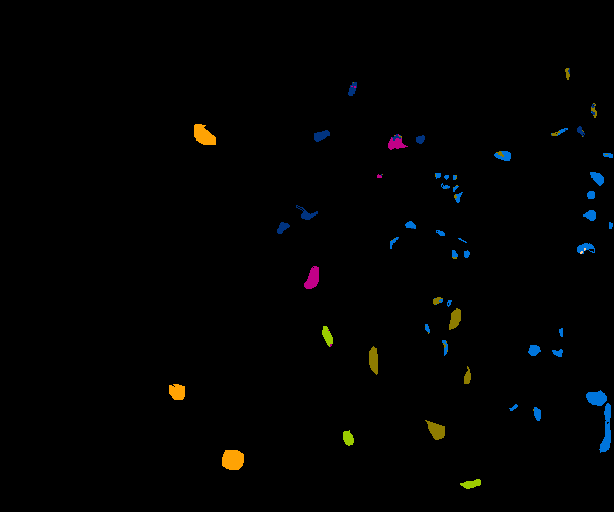}}
	\caption{Ground truth and segmentations of \ac{mnf}-transformed \ac{ksc} data produced by the competing methods.}
	\label{fig:ksc-mnf-gt-seg}
\end{figure*}

\begin{figure*}[ht]
    \centering
    \subfloat[Ground truth]{\label{img:pu-raw-gt}\includegraphics[width=0.22\textwidth]{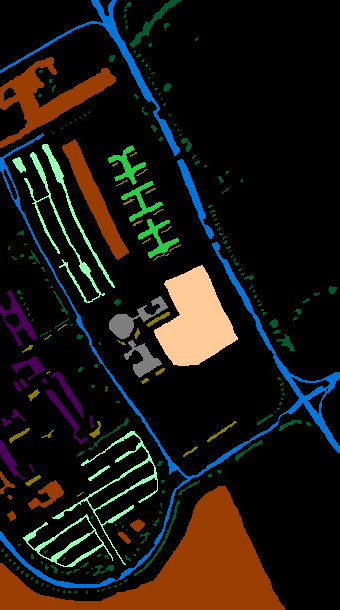}}
    \quad
    \subfloat[$k$-means (0.534)]{\label{img:pu-raw-kmeans-seg}\includegraphics[width=0.22\textwidth]{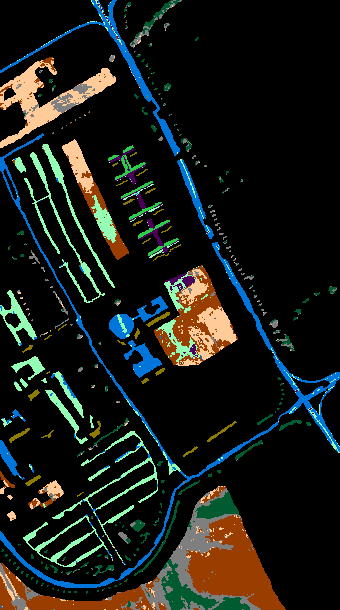}}
    \quad
    \subfloat[GMM (0.517)]{\label{img:pu-raw-gmm-seg}\includegraphics[width=0.22\textwidth]{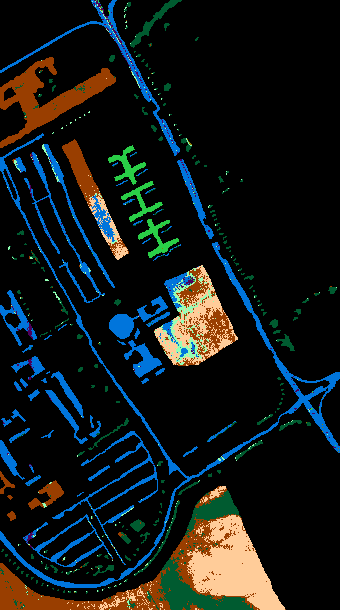}}
    \quad
    \subfloat[BGM (0.444)]{\label{img:pu-raw-bgm-seg}\includegraphics[width=0.22\textwidth]{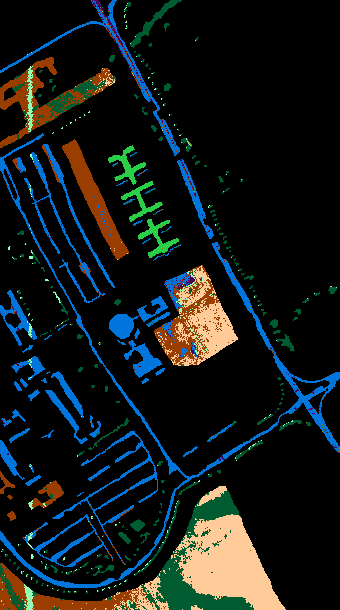}}
    \\
    \subfloat[3D-CAE (0.535)]{\label{img:pu-raw-3dcae-seg}\includegraphics[width=0.22\textwidth]{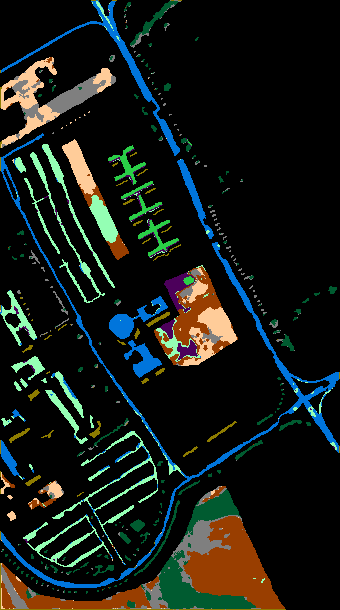}}
    \quad
    \subfloat[MS-2 (0.564)]{\label{img:pu-raw-2norm-seg}\includegraphics[width=0.22\textwidth]{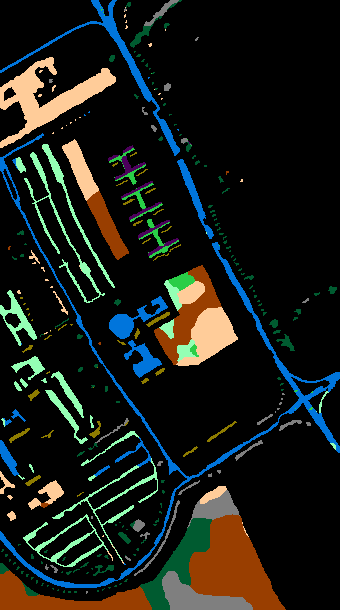}}
    \quad
    \subfloat[Ours (0.562)]{\label{img:pu-raw-an2norm-seg}\includegraphics[width=0.22\textwidth]{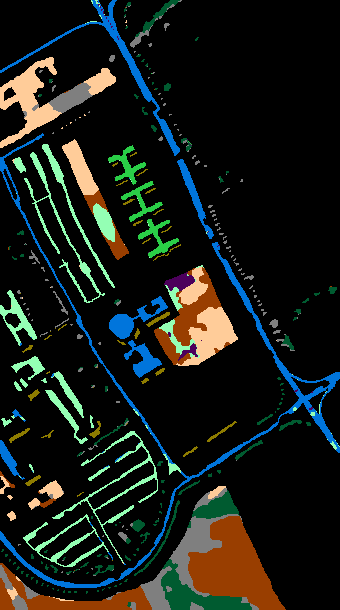}}
    \caption{Ground truth and segmentations of raw \ac{pu} data produced by the competing methods.}
    \label{fig:pu-raw-gt-seg}
\end{figure*}

\begin{figure*}[ht]
    \centering
    \subfloat[Ground truth]{\label{img:pu-mnf-gt}\includegraphics[width=0.22\textwidth]{input/pu_gt.png}}
    \quad
    \subfloat[$k$-means (0.512)]{\label{img:pu-mnf-kmeans-seg}\includegraphics[width=0.22\textwidth]{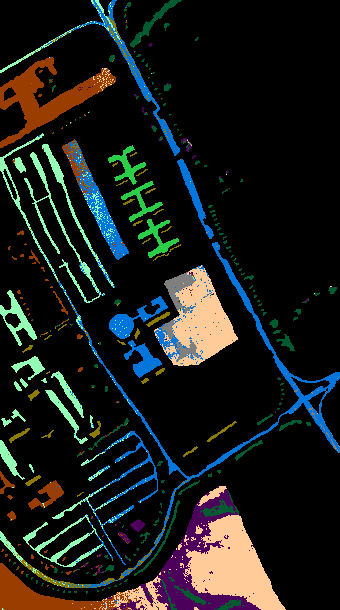}}
    \quad
    \subfloat[GMM (0.525)]{\label{img:pu-mnf-gmm-seg}\includegraphics[width=0.22\textwidth]{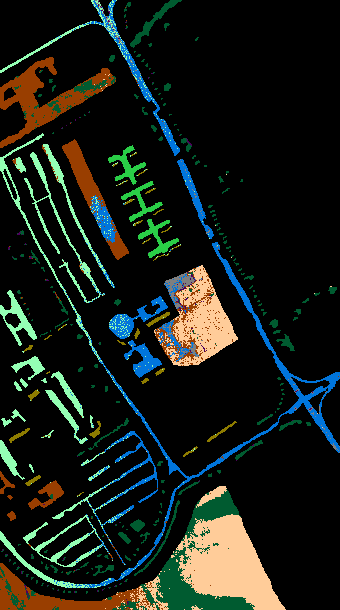}}
    \quad
    \subfloat[BGM (0.475)]{\label{img:pu-mnf-bgm-seg}\includegraphics[width=0.22\textwidth]{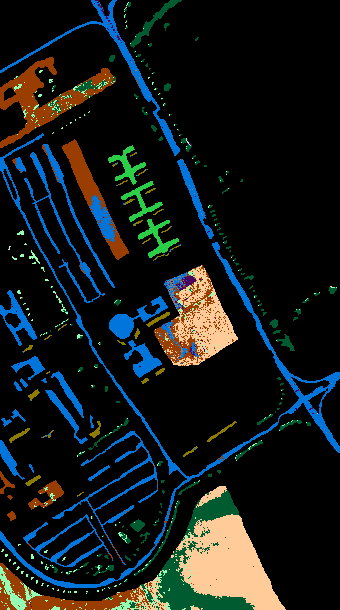}}
    \\
    \subfloat[3D-CAE (0.500)]{\label{img:pu-mnf-3dcae-seg}\includegraphics[width=0.22\textwidth]{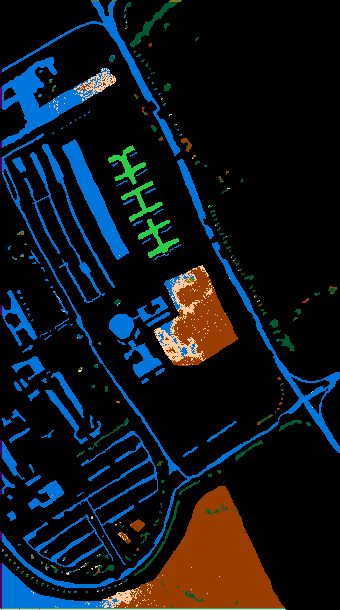}}
    \quad
    \subfloat[MS-2 (0.518)]{\label{img:pu-mnf-2norm-seg}\includegraphics[width=0.22\textwidth]{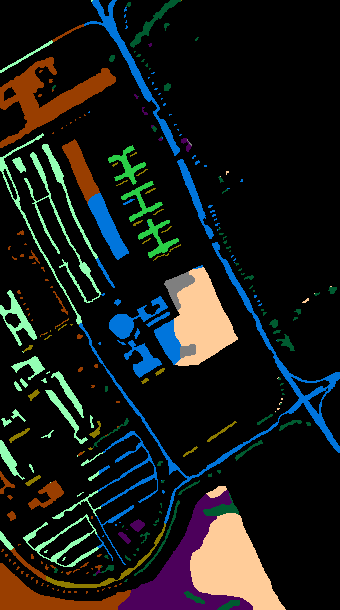}}
    \quad
    \subfloat[Ours (0.561)]{\label{img:pu-mnf-an2norm-seg}\includegraphics[width=0.22\textwidth]{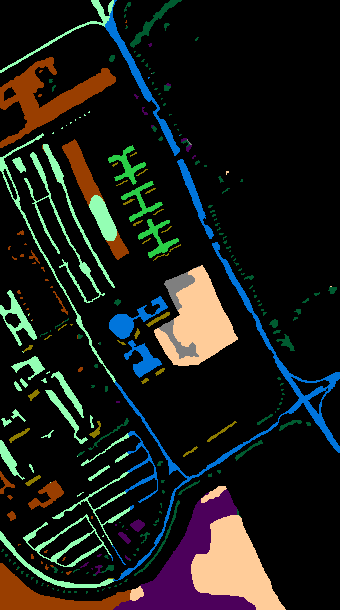}}
    \caption{Ground truth and segmentations of \ac{mnf}-transformed \ac{pu} data produced by the competing methods.}
    \label{fig:pu-mnf-gt-seg}
\end{figure*}

\begin{figure*}[ht]
    \centering
    \subfloat[Ground truth]{\label{img:pu-3dcae-fl-gt}\includegraphics[width=0.22\textwidth]{input/pu_gt.png}}
    \quad
    \subfloat[$k$-means (0.603)]{\label{img:pu-3dcae-fl-kmeans-seg}\includegraphics[width=0.22\textwidth]{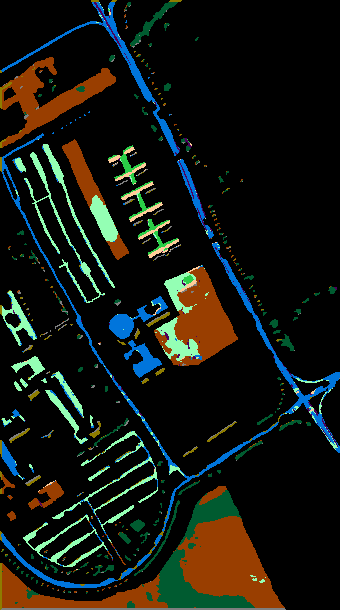}}
    \quad
    \subfloat[GMM (0.580)]{\label{img:pu-3dcae-fl-gmm-seg}\includegraphics[width=0.22\textwidth]{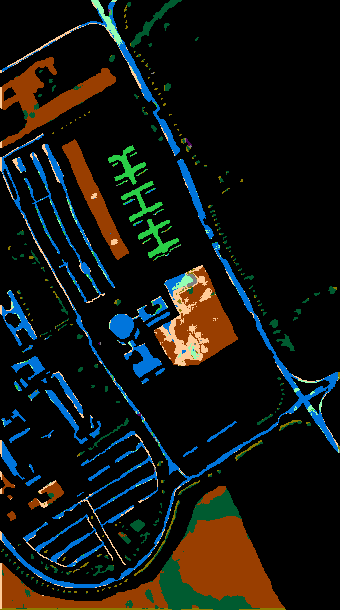}}
    \quad
    \subfloat[BGM (0.571)]{\label{img:pu-3dcae-fl-bgm-seg}\includegraphics[width=0.22\textwidth]{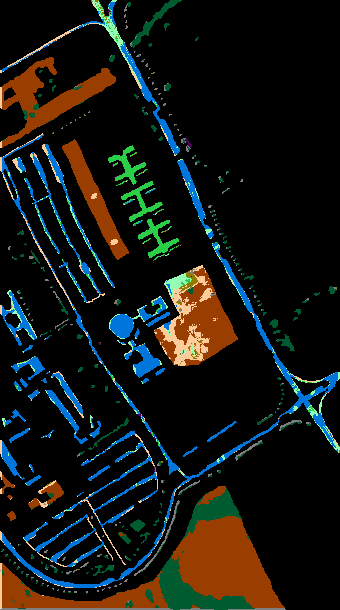}}
    \\
    \subfloat[3D-CAE (0.481)]{\label{img:pu-3dcae-fl-3dcae-seg}\includegraphics[width=0.22\textwidth]{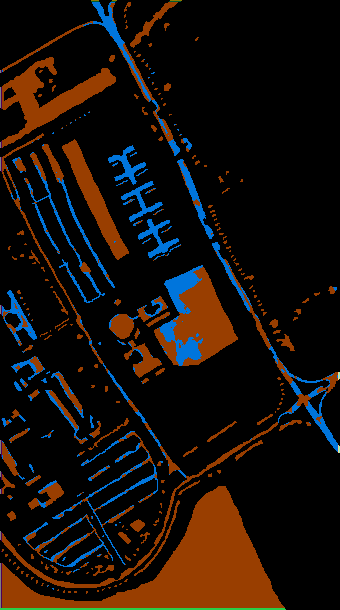}}
    \quad
    \subfloat[MS-2 (0.576)]{\label{img:pu-3dcae-fl-2norm-seg}\includegraphics[width=0.22\textwidth]{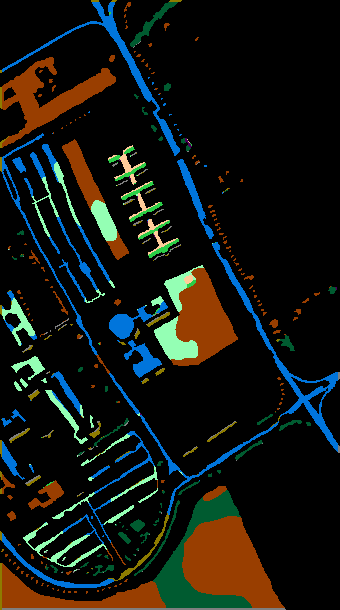}}
    \quad
    \subfloat[Ours (0.636)]{\label{img:pu-3dcae-fl-an2norm-seg}\includegraphics[width=0.22\textwidth]{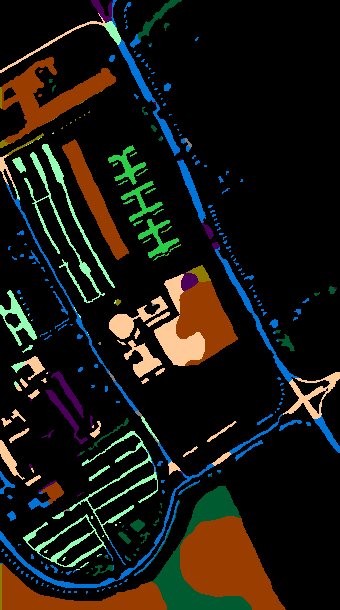}}
    \caption{Ground truth and segmentations of the learned \ac{3dfeats} features of the \ac{pu} data produced by the competing methods.}
    \label{fig:pu-3dcae-fl-gt-seg}
\end{figure*}

\begin{figure*}[ht]
    \centering
    \subfloat[Ground truth]{\label{img:sa-raw-gt}\includegraphics[width=0.22\textwidth]{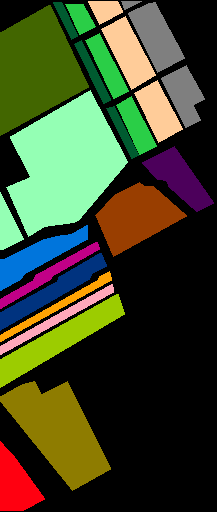}}
    \quad
    \subfloat[$k$-means (0.669)]{\label{img:sa-raw-kmeans-seg}\includegraphics[width=0.22\textwidth]{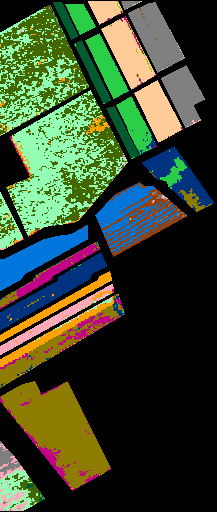}}
    \quad
    \subfloat[GMM (0.697)]{\label{img:sa-raw-gmm-seg}\includegraphics[width=0.22\textwidth]{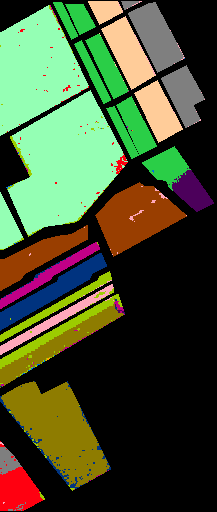}}
    \quad
    \subfloat[BGM (0.676)]{\label{img:sa-raw-bgm-seg}\includegraphics[width=0.22\textwidth]{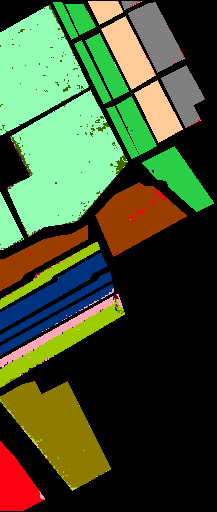}}
    \\
    \subfloat[3D-CAE (0.597)]{\label{img:sa-raw-3dcae-seg}\includegraphics[width=0.22\textwidth]{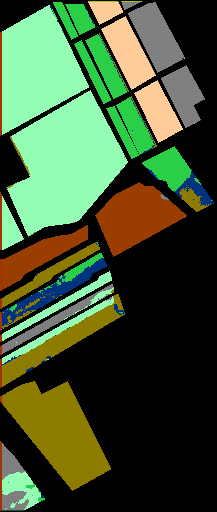}}
    \quad
    \subfloat[MS-2 (0.733)]{\label{img:sa-raw-2norm-seg}\includegraphics[width=0.22\textwidth]{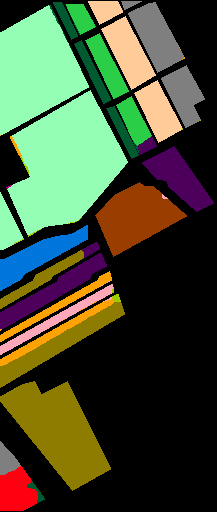}}
    \quad
    \subfloat[Ours (0.810)]{\label{img:sa-raw-an2norm-seg}\includegraphics[width=0.22\textwidth]{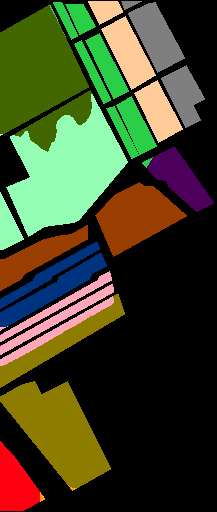}}
    \caption{Ground truth and segmentations of raw \ac{sa} data produced by the competing methods.}
    \label{fig:sa-raw-gt-seg}
\end{figure*}

\begin{figure*}[ht]
    \centering
    \subfloat[Ground truth]{\label{img:sa-mnf-gt}\includegraphics[width=0.22\textwidth]{input/sa_gt.png}}
    \quad
    \subfloat[$k$-means (0.731)]{\label{img:sa-mnf-kmeans-seg}\includegraphics[width=0.22\textwidth]{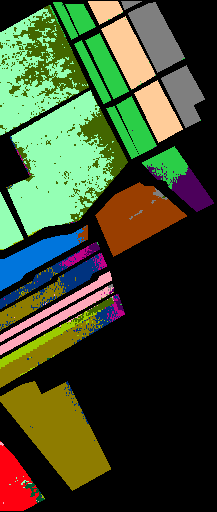}}
    \quad
    \subfloat[GMM (0.712)]{\label{img:sa-mnf-gmm-seg}\includegraphics[width=0.22\textwidth]{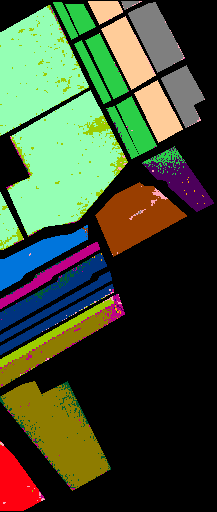}}
    \quad
    \subfloat[BGM (0.724)]{\label{img:sa-mnf-bgm-seg}\includegraphics[width=0.22\textwidth]{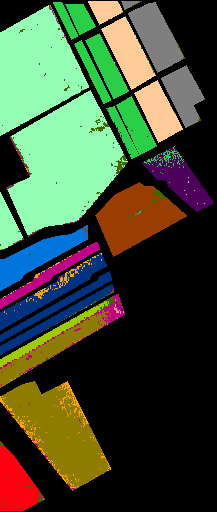}}
    \\
    \subfloat[3D-CAE (0.639)]{\label{img:sa-mnf-3dcae-seg}\includegraphics[width=0.22\textwidth]{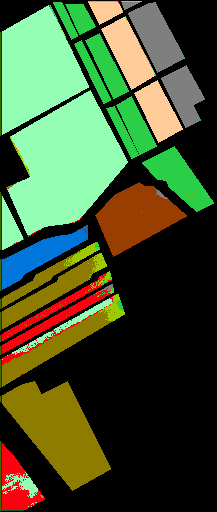}}
    \quad
    \subfloat[MS-2 (0.764)]{\label{img:sa-mnf-2norm-seg}\includegraphics[width=0.22\textwidth]{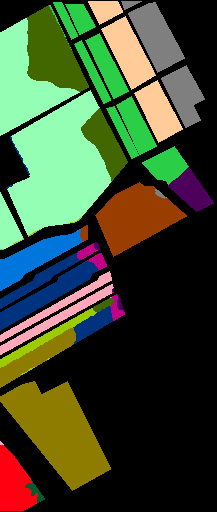}}
    \quad
    \subfloat[Ours (0.896)]{\label{img:sa-mnf-an2norm-seg}\includegraphics[width=0.22\textwidth]{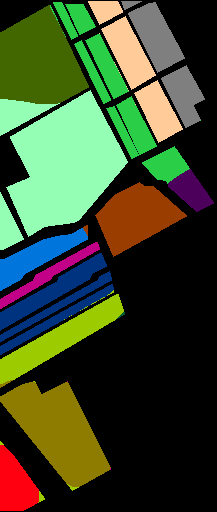}}
    \caption{Ground truth and segmentations of \ac{mnf}-transformed \ac{sa} data produced by the competing methods.}
    \label{fig:sa-mnf-gt-seg}
\end{figure*}

\begin{figure*}[ht]
    \centering
    \subfloat[Ground truth]{\label{img:sa-3dcae-fl-gt}\includegraphics[width=0.22\textwidth]{input/sa_gt.png}}
    \quad
    \subfloat[$k$-means (0.657)]{\label{img:sa-3dcae-fl-kmeans-seg}\includegraphics[width=0.22\textwidth]{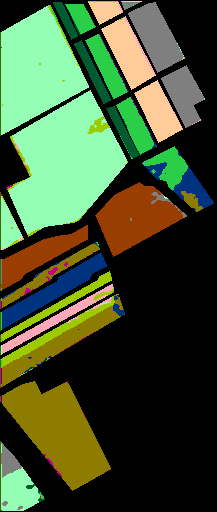}}
    \quad
    \subfloat[GMM (0.617)]{\label{img:sa-3dcae-fl-gmm-seg}\includegraphics[width=0.22\textwidth]{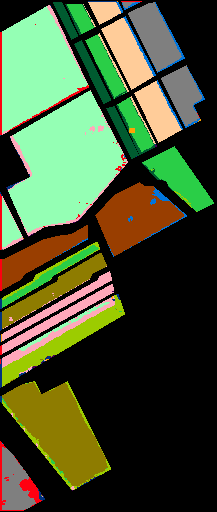}}
    \quad
    \subfloat[BGM (0.628)]{\label{img:sa-3dcae-fl-bgm-seg}\includegraphics[width=0.22\textwidth]{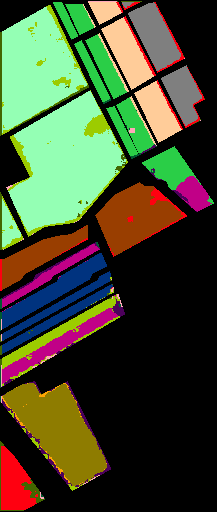}}
    \\
    \subfloat[3D-CAE (0.452)]{\label{img:sa-3dcae-fl-3dcae-seg}\includegraphics[width=0.22\textwidth]{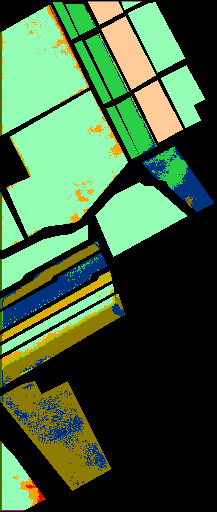}}
    \quad
    \subfloat[MS-2 (0.688)]{\label{img:sa-3dcae-fl-2norm-seg}\includegraphics[width=0.22\textwidth]{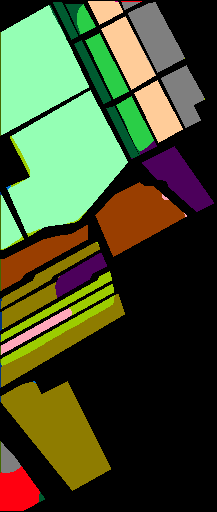}}
    \quad
    \subfloat[Ours (0.662)]{\label{img:sa-3dcae-fl-an2norm-seg}\includegraphics[width=0.22\textwidth]{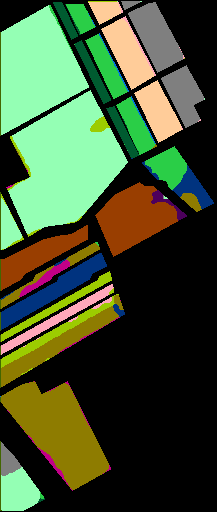}}
    \caption{Ground truth and segmentations of the learned \ac{3dfeats} features of the \ac{sa} data produced by the competing methods.}
    \label{fig:sa-3dcae-fl-gt-seg}
\end{figure*}

\noindent
The achieved scores are given in \cref{tab:results-oa,tab:results-aa,tab:results-kc}.
\emph{Std} describes the standard deviation of the 10 trials with different seeds.
Its rows have a gray background for easier distinction.
\emph{Fixed} refers to the fixed seed with which $\lambda$ and $\epsilon$ were tuned.
The best mean result and the highest maximum score for each dataset variant are highlighted in bold letters and with a green table cell, respectively.
Please note that the authors of SSC run their algorithm ten times on every dataset with random initialization and report the best result.
Therefore, we compare our maximum with their scores.
They found the best parameters for their model using cross validation with the \ac{oa} score.
The authors of NLTV do not say which score is reported.
We therefore list it as \emph{mean} in the tables and compare it with the mean of the other methods.
The segmentations are shown in \cref{fig:ip-raw-gt-seg,fig:ip-mnf-gt-seg,fig:ip-3dcae-fl-gt-seg,fig:ksc-raw-gt-seg,fig:ksc-mnf-gt-seg,fig:pu-raw-gt-seg,fig:pu-mnf-gt-seg,fig:pu-3dcae-fl-gt-seg,fig:sa-raw-gt-seg,fig:sa-mnf-gt-seg,fig:sa-3dcae-fl-gt-seg}.
We show the segmentations that correspond to the seeds yielding the fifth best overall accuracy out of ten.
The scores are given in parentheses.
The colors used for visualization are adapted from \cite{Gr10}.
Pixels that are unlabeled in the ground truth are marked in black.
\begin{table*}[ht]
	\centering
	\adjustbox{max width=2\columnwidth}{\begin{tabular}{ll|ccc|cc|ccc|ccc|}
\toprule
 & Image & \multicolumn{3}{|c|}{\textbf{\ac{ip}}} & \multicolumn{2}{|c|}{\textbf{\ac{ksc}}} & \multicolumn{3}{|c|}{\textbf{\ac{pu}}} & \multicolumn{3}{|c|}{\textbf{\ac{sa}}} \\
 & Type & raw & MNF & \ac{3dfeats} & raw & MNF & raw & MNF & \ac{3dfeats} & raw & MNF & \ac{3dfeats} \\
Method & Stats &  &  &  &  &  &  &  &  &  &  &  \\
\midrule
\multirow[c]{4}{*}{$k$\textbf{-means}} & Min & 0.360 & 0.485 & 0.393 & 0.178 & 0.616 & 0.534 & 0.509 & 0.603 & 0.669 & 0.707 & 0.636 \\
 & Mean & 0.368 & 0.502 & 0.402 & 0.178 & 0.624 & 0.534 & 0.515 & 0.604 & 0.669 & 0.723 & 0.654 \\
 & Max & 0.378 & 0.506 & 0.410 & 0.178 & 0.632 & 0.534 & 0.550 & 0.611 & 0.671 & 0.732 & 0.663 \\
 \rowcolor{std-color} & Std & 0.007 & 0.006 & 0.005 & 0.000 & 0.005 & 0.000 & 0.012 & 0.002 & 0.001 & 0.011 & 0.009 \\
\cline{1-13}
\multirow[c]{4}{*}{\textbf{\ac{gmm}}} & Min & 0.376 & 0.489 & 0.447 & 0.178 & 0.591 & 0.499 & 0.495 & 0.576 & 0.666 & 0.683 & 0.591 \\
 & Mean & 0.402 & 0.514 & 0.467 & 0.178 & 0.612 & 0.528 & 0.515 & 0.580 & 0.691 & 0.715 & 0.625 \\
 & Max & 0.438 & 0.577 & 0.493 & 0.179 & 0.646 & 0.556 & 0.528 & 0.585 & 0.715 & 0.753 & 0.655 \\
 \rowcolor{std-color} & Std & 0.021 & 0.024 & 0.018 & 0.000 & 0.017 & 0.025 & 0.013 & 0.004 & 0.018 & 0.022 & 0.020 \\
\cline{1-13}
\multirow[c]{4}{*}{\textbf{\ac{bgm}}} & Min & 0.379 & 0.488 & 0.357 & 0.178 & 0.580 & 0.443 & 0.463 & 0.565 & 0.668 & 0.696 & 0.572 \\
 & Mean & 0.404 & 0.514 & 0.402 & 0.178 & 0.612 & 0.493 & 0.477 & 0.569 & 0.677 & 0.721 & 0.632 \\
 & Max & 0.438 & 0.590 & 0.454 & 0.179 & 0.626 & 0.569 & 0.518 & 0.571 & 0.693 & 0.748 & 0.685 \\
 \rowcolor{std-color} & Std & 0.019 & 0.028 & 0.034 & 0.000 & 0.014 & 0.065 & 0.016 & 0.003 & 0.008 & 0.017 & 0.040 \\
\cline{1-13}
\multirow[c]{4}{*}{\textbf{\ac{3dcae}}} & Min & 0.390 & 0.456 & 0.240 & 0.185 & 0.330 & 0.460 & 0.405 & 0.363 & 0.584 & 0.590 & 0.415 \\
 & Mean & 0.414 & 0.473 & 0.391 & 0.213 & 0.446 & 0.540 & 0.477 & 0.468 & 0.595 & 0.625 & 0.444 \\
 & Max & 0.436 & 0.486 & 0.426 & \cellcolor{highest-max-color}0.309 & 0.600 & 0.599 & 0.561 & 0.542 & 0.601 & 0.657 & 0.479 \\
 \rowcolor{std-color} & Std & 0.015 & 0.009 & 0.055 & 0.051 & 0.069 & 0.044 & 0.055 & 0.047 & 0.005 & 0.026 & 0.020 \\
\cline{1-13}
\textbf{SSC} & Max & 0.429 & - & - & - & - & \cellcolor{highest-max-color}0.682 & - & - & \cellcolor{highest-max-color}0.818 & - & - \\
\cline{1-13}
\textbf{NLTV} & Mean & 0.424 & - & - & \textbf{0.415} & - & 0.440 & - & - & - & - & - \\
\cline{1-13}
\multirow[c]{5}{*}{\textbf{MS-2}} & Min & 0.475 & 0.553 & 0.442 & 0.179 & 0.676 & 0.564 & 0.518 & 0.576 & 0.733 & 0.713 & 0.655 \\
 & Mean & 0.497 & 0.588 & 0.467 & 0.179 & \textbf{0.676} & \textbf{0.564} & 0.525 & 0.576 & 0.733 & 0.750 & \textbf{0.683} \\
 & Max & 0.541 & 0.631 & 0.482 & 0.179 & \cellcolor{highest-max-color}0.676 & 0.564 & 0.593 & 0.578 & 0.733 & 0.764 & \cellcolor{highest-max-color}0.709 \\
 \rowcolor{std-color} & Std & 0.018 & 0.038 & 0.018 & 0.000 & 0.000 & 0.000 & 0.024 & 0.000 & 0.000 & 0.019 & 0.019 \\
 & Fixed & 0.500 & 0.629 & 0.482 & 0.179 & 0.650 & 0.564 & 0.593 & 0.576 & 0.733 & 0.764 & 0.688 \\
\cline{1-13}
\multirow[c]{5}{*}{\textbf{Ours}} & Min & 0.505 & 0.582 & 0.482 & 0.179 & 0.666 & 0.562 & 0.560 & 0.605 & 0.807 & 0.723 & 0.646 \\
 & Mean & \textbf{0.554} & \textbf{0.671} & \textbf{0.494} & 0.180 & 0.668 & 0.562 & \textbf{0.567} & \textbf{0.633} & \textbf{0.809} & \textbf{0.828} & 0.667 \\
 & Max & \cellcolor{highest-max-color}0.639 & \cellcolor{highest-max-color}0.697 & \cellcolor{highest-max-color}0.499 & 0.180 & 0.669 & 0.562 & \cellcolor{highest-max-color}0.615 & \cellcolor{highest-max-color}0.636 & 0.810 & \cellcolor{highest-max-color}0.896 & 0.690 \\
 \rowcolor{std-color} & Std & 0.051 & 0.034 & 0.006 & 0.000 & 0.001 & 0.000 & 0.017 & 0.010 & 0.001 & 0.074 & 0.016 \\
 & Fixed & 0.571 & 0.697 & 0.523 & 0.179 & 0.716 & 0.562 & 0.615 & 0.636 & 0.810 & 0.896 & 0.689 \\
\cline{1-13}
\bottomrule
\end{tabular}
}
	\caption{Achieved \ac{oa} of the competing methods on the four datasets.
	}
	\label{tab:results-oa}
\end{table*}
\begin{table*}[ht]
	\centering
	\adjustbox{max width=2\columnwidth}{\begin{tabular}{ll|ccc|cc|ccc|ccc|}
\toprule
 & Image & \multicolumn{3}{|c|}{\textbf{\ac{ip}}} & \multicolumn{2}{|c|}{\textbf{\ac{ksc}}} & \multicolumn{3}{|c|}{\textbf{\ac{pu}}} & \multicolumn{3}{|c|}{\textbf{\ac{sa}}} \\
 & Type & raw & MNF & \ac{3dfeats} & raw & MNF & raw & MNF & \ac{3dfeats} & raw & MNF & \ac{3dfeats} \\
Method & Stats &  &  &  &  &  &  &  &  &  &  &  \\
\midrule
\multirow[c]{4}{*}{$k$-\textbf{means}} & Min & 0.375 & 0.457 & 0.268 & 0.079 & 0.405 & 0.526 & 0.582 & 0.474 & 0.687 & 0.631 & 0.565 \\
 & Mean & 0.404 & 0.464 & 0.291 & 0.079 & 0.415 & 0.526 & 0.585 & 0.478 & 0.687 & 0.644 & \textbf{0.626} \\
 & Max & 0.440 & 0.473 & 0.310 & 0.079 & 0.441 & 0.526 & 0.598 & \cellcolor{highest-max-color}0.505 & 0.688 & 0.653 & 0.656 \\
 \rowcolor{std-color} & Std & 0.027 & 0.005 & 0.019 & 0.000 & 0.011 & 0.000 & 0.005 & 0.010 & 0.000 & 0.009 & 0.035 \\
\cline{1-13}
\multirow[c]{4}{*}{\textbf{\ac{gmm}}} & Min & 0.361 & 0.481 & 0.301 & 0.078 & 0.396 & 0.420 & 0.466 & 0.398 & 0.547 & 0.609 & 0.461 \\
 & Mean & 0.386 & 0.531 & \textbf{0.343} & 0.079 & 0.410 & 0.483 & 0.528 & 0.404 & 0.615 & 0.654 & 0.527 \\
 & Max & 0.418 & 0.595 & \cellcolor{highest-max-color}0.396 & 0.079 & 0.453 & 0.546 & 0.587 & 0.412 & 0.650 & 0.695 & 0.585 \\
 \rowcolor{std-color} & Std & 0.021 & 0.036 & 0.030 & 0.000 & 0.019 & 0.044 & 0.058 & 0.006 & 0.041 & 0.029 & 0.037 \\
\cline{1-13}
\multirow[c]{4}{*}{\textbf{\ac{bgm}}} & Min & 0.357 & 0.469 & 0.216 & 0.078 & 0.393 & 0.406 & 0.429 & 0.383 & 0.536 & 0.615 & 0.533 \\
 & Mean & 0.378 & 0.531 & 0.264 & 0.079 & 0.422 & 0.409 & 0.480 & 0.393 & 0.575 & 0.662 & 0.566 \\
 & Max & 0.405 & 0.603 & 0.340 & 0.079 & 0.436 & 0.412 & 0.582 & 0.398 & 0.604 & 0.688 & 0.611 \\
 \rowcolor{std-color} & Std & 0.017 & 0.041 & 0.042 & 0.000 & 0.017 & 0.003 & 0.051 & 0.007 & 0.028 & 0.024 & 0.030 \\
\cline{1-13}
\multirow[c]{4}{*}{\textbf{\ac{3dcae}}} & Min & 0.191 & 0.317 & 0.088 & 0.101 & 0.201 & 0.554 & 0.306 & 0.148 & 0.396 & 0.405 & 0.251 \\
 & Mean & 0.224 & 0.355 & 0.196 & \textbf{0.115} & 0.274 & 0.573 & 0.356 & 0.209 & 0.410 & 0.473 & 0.303 \\
 & Max & 0.255 & 0.409 & 0.237 & \cellcolor{highest-max-color}0.161 & 0.388 & 0.598 & 0.405 & 0.297 & 0.419 & 0.540 & 0.363 \\
 \rowcolor{std-color} & Std & 0.019 & 0.032 & 0.042 & 0.024 & 0.056 & 0.018 & 0.036 & 0.052 & 0.007 & 0.039 & 0.035 \\
\cline{1-13}
\multirow[c]{5}{*}{\textbf{MS-2}} & Min & 0.405 & 0.529 & 0.300 & 0.078 & 0.459 & 0.539 & 0.548 & 0.424 & 0.724 & 0.631 & 0.539 \\
 & Mean & 0.471 & 0.540 & 0.309 & 0.078 & \textbf{0.459} & 0.539 & 0.552 & 0.426 & \textbf{0.724} & 0.671 & 0.590 \\
 & Max & 0.527 & 0.552 & 0.360 & 0.079 & \cellcolor{highest-max-color}0.459 & 0.539 & 0.588 & 0.440 & \cellcolor{highest-max-color}0.724 & 0.685 & 0.648 \\
 \rowcolor{std-color} & Std & 0.042 & 0.010 & 0.019 & 0.000 & 0.000 & 0.000 & 0.013 & 0.005 & 0.000 & 0.021 & 0.039 \\
 & Fixed & 0.449 & 0.488 & 0.360 & 0.079 & 0.436 & 0.539 & 0.588 & 0.424 & 0.724 & 0.685 & 0.597 \\
\cline{1-13}
\multirow[c]{5}{*}{\textbf{Ours}} & Min & 0.497 & 0.538 & 0.293 & 0.078 & 0.454 & 0.601 & 0.613 & 0.452 & 0.671 & 0.639 & 0.568 \\
 & Mean & \textbf{0.541} & \textbf{0.607} & 0.331 & 0.079 & 0.454 & \textbf{0.601} & \textbf{0.616} & \textbf{0.491} & 0.675 & \textbf{0.730} & 0.614 \\
 & Max & \cellcolor{highest-max-color}0.585 & \cellcolor{highest-max-color}0.633 & 0.355 & 0.079 & 0.455 & \cellcolor{highest-max-color}0.601 & \cellcolor{highest-max-color}0.636 & 0.495 & 0.676 & \cellcolor{highest-max-color}0.768 & \cellcolor{highest-max-color}0.676 \\
 \rowcolor{std-color} & Std & 0.028 & 0.030 & 0.018 & 0.000 & 0.001 & 0.000 & 0.007 & 0.014 & 0.002 & 0.045 & 0.039 \\
 & Fixed & 0.539 & 0.607 & 0.392 & 0.078 & 0.530 & 0.601 & 0.636 & 0.495 & 0.676 & 0.768 & 0.629 \\
\cline{1-13}
\bottomrule
\end{tabular}
}
	\caption{Achieved \ac{aa} of the competing methods on the four datasets.
	}
	\label{tab:results-aa}
\end{table*}
\begin{table*}[ht]
	\centering
	\adjustbox{max width=2\columnwidth}{\begin{tabular}{ll|ccc|cc|ccc|ccc|}
\toprule
 & Image & \multicolumn{3}{|c|}{\textbf{\ac{ip}}} & \multicolumn{2}{|c|}{\textbf{\ac{ksc}}} & \multicolumn{3}{|c|}{\textbf{\ac{pu}}} & \multicolumn{3}{|c|}{\textbf{\ac{sa}}} \\
 & Type & raw & MNF & \ac{3dfeats} & raw & MNF & raw & MNF & \ac{3dfeats} & raw & MNF & \ac{3dfeats} \\
Method & Stats &  &  &  &  &  &  &  &  &  &  &  \\
\midrule
\multirow[c]{4}{*}{$k$-\textbf{means}} & Min & 0.289 & 0.426 & 0.314 & 0.002 & 0.560 & 0.434 & 0.421 & 0.482 & 0.632 & 0.669 & 0.590 \\
 & Mean & 0.298 & 0.442 & 0.319 & 0.002 & 0.570 & 0.434 & 0.428 & 0.484 & 0.632 & 0.689 & 0.610 \\
 & Max & 0.307 & 0.446 & 0.321 & 0.003 & 0.579 & 0.434 & 0.467 & 0.493 & 0.634 & 0.699 & 0.621 \\
 \rowcolor{std-color} & Std & 0.008 & 0.006 & 0.002 & 0.000 & 0.006 & 0.000 & 0.014 & 0.003 & 0.001 & 0.013 & 0.011 \\
\cline{1-13}
\multirow[c]{4}{*}{\textbf{\ac{gmm}}} & Min & 0.305 & 0.439 & 0.359 & 0.001 & 0.531 & 0.383 & 0.400 & 0.437 & 0.623 & 0.644 & 0.538 \\
 & Mean & 0.328 & 0.464 & 0.381 & 0.002 & 0.558 & 0.407 & 0.425 & 0.444 & 0.651 & 0.680 & 0.577 \\
 & Max & 0.360 & 0.521 & 0.415 & 0.003 & 0.597 & 0.424 & 0.437 & 0.451 & 0.679 & 0.722 & 0.611 \\
 \rowcolor{std-color} & Std & 0.020 & 0.023 & 0.023 & 0.001 & 0.020 & 0.016 & 0.013 & 0.006 & 0.021 & 0.024 & 0.023 \\
\cline{1-13}
\multirow[c]{4}{*}{\textbf{\ac{bgm}}} & Min & 0.309 & 0.439 & 0.250 & 0.001 & 0.525 & 0.320 & 0.356 & 0.426 & 0.626 & 0.658 & 0.529 \\
 & Mean & 0.330 & 0.463 & 0.306 & 0.002 & 0.558 & 0.364 & 0.373 & 0.432 & 0.635 & 0.686 & 0.590 \\
 & Max & 0.358 & 0.534 & 0.371 & 0.003 & 0.573 & 0.430 & 0.420 & 0.435 & 0.654 & 0.717 & 0.650 \\
 \rowcolor{std-color} & Std & 0.016 & 0.026 & 0.044 & 0.000 & 0.015 & 0.057 & 0.020 & 0.005 & 0.010 & 0.019 & 0.042 \\
\cline{1-13}
\multirow[c]{4}{*}{\textbf{\ac{3dcae}}} & Min & 0.270 & 0.377 & 0.019 & 0.049 & 0.228 & 0.373 & 0.105 & 0.159 & 0.524 & 0.535 & 0.314 \\
 & Mean & 0.302 & 0.395 & 0.258 & \textbf{0.080} & 0.366 & 0.446 & 0.296 & 0.217 & 0.536 & 0.573 & 0.347 \\
 & Max & 0.335 & 0.416 & 0.305 & \cellcolor{highest-max-color}0.191 & 0.543 & 0.501 & 0.377 & 0.363 & 0.543 & 0.612 & 0.391 \\
 \rowcolor{std-color} & Std & 0.021 & 0.013 & 0.086 & 0.058 & 0.080 & 0.041 & 0.082 & 0.057 & 0.006 & 0.030 & 0.023 \\
\cline{1-13}
\textbf{SSC} & Max & 0.365 & - & - & - & - & \cellcolor{highest-max-color}0.622 & - & - & \cellcolor{highest-max-color}0.798 & - & - \\
\cline{1-13}
\multirow[c]{5}{*}{\textbf{MS-2}} & Min & 0.405 & 0.501 & 0.375 & 0.002 & 0.630 & 0.465 & 0.423 & 0.436 & 0.698 & 0.675 & 0.606 \\
 & Mean & 0.431 & 0.537 & 0.386 & 0.002 & \textbf{0.630} & 0.465 & 0.432 & 0.437 & 0.698 & 0.719 & \textbf{0.641} \\
 & Max & 0.487 & 0.580 & 0.399 & 0.003 & \cellcolor{highest-max-color}0.630 & 0.466 & 0.509 & 0.437 & 0.698 & 0.735 & \cellcolor{highest-max-color}0.671 \\
 \rowcolor{std-color} & Std & 0.022 & 0.038 & 0.009 & 0.000 & 0.000 & 0.000 & 0.027 & 0.000 & 0.000 & 0.023 & 0.023 \\
 & Fixed & 0.428 & 0.577 & 0.392 & 0.003 & 0.603 & 0.466 & 0.509 & 0.437 & 0.698 & 0.735 & 0.646 \\
\cline{1-13}
\multirow[c]{5}{*}{\textbf{Ours}} & Min & 0.445 & 0.531 & 0.392 & 0.002 & 0.617 & 0.472 & 0.479 & 0.485 & 0.785 & 0.689 & 0.601 \\
 & Mean & \textbf{0.501} & \textbf{0.631} & \textbf{0.413} & 0.003 & 0.619 & \textbf{0.472} & \textbf{0.486} & \textbf{0.519} & \textbf{0.787} & \textbf{0.807} & 0.625 \\
 & Max & \cellcolor{highest-max-color}0.587 & \cellcolor{highest-max-color}0.659 & \cellcolor{highest-max-color}0.426 & 0.003 & 0.620 & 0.472 & \cellcolor{highest-max-color}0.539 & \cellcolor{highest-max-color}0.523 & 0.788 & \cellcolor{highest-max-color}0.884 & 0.650 \\
 \rowcolor{std-color} & Std & 0.051 & 0.037 & 0.010 & 0.000 & 0.001 & 0.000 & 0.019 & 0.012 & 0.001 & 0.084 & 0.018 \\
 & Fixed & 0.511 & 0.659 & 0.446 & 0.002 & 0.680 & 0.472 & 0.539 & 0.523 & 0.788 & 0.884 & 0.648 \\
\cline{1-13}
\bottomrule
\end{tabular}
}
	\caption{Achieved \ac{kc} of the competing methods on the four datasets.
	}
	\label{tab:results-kc}
\end{table*}
All three metrics show that our framework (including the \ac{mnf}) achieves very good scores on all datasets.
This shows that our indicator function indeed can handle the spectral variability.
Furthermore, the \ac{mnf} transform and the square root in the indicator function enable the model to deal with significant spectral outliers, which are common in hyperspectral datasets.
Finally, the regularization of the perimeter leads to very homogeneous segmentations and makes it possible to disentangle overlapping clusters.

\subsubsection{\acl{ip}}
On \ac{ip}, the \ac{ms} functional with our proposed indicator function beats all other methods significantly.
In case of \ac{oa} and \ac{kc}, the achieved mean score is for all dataset types even better than the maximum scores of the other methods.
However, our method shows the largest fluctuation.
The dataset itself contains many spectrally similar segments and is therefore very challenging.

\subsubsection{\acl{ksc}}
Considering \ac{ksc}, we see that the performance on the raw dataset is bad for all methods.
Merely NLTV is able to find a segmentation that is not completely useless although the score means that not even half of the pixels are correctly classified by the algorithm.
The \ac{ms}-based methods (MS-2 and ours) use the $k$-means result as initialization.
We can observe that the algorithm is not able to recover from the low quality initialization that it obtains from $k$-means since the objective function is highly non-convex.
An application of the \ac{mnf} before segmenting the image helps all methods to significantly improve the results.
Although MS-2 achieves the best result on \ac{ksc}, we point out that the score we obtained with our modification using the fixed seed is higher than the best score of MS-2.
This means that our functional is able to deliver a better description of the data but gets stuck in a local minimum for all of the ten randomly sampled seeds.
\ac{3dcae} shows a very high fluctuation in the segmentation results.

\subsubsection{\acl{pu}}
The \ac{pu} dataset is the only one that was not captured with an \ac{aviris} sensor.
SSC achieves the best score for this data.
Neither \ac{mnf} nor \ac{3dfeats} help to beat SSC although both give a performance boost for the maximum score.
The result indicates that a graph-based spectral clustering approach like SSC is more suitable for \ac{pu} than the others.
Also here, the fluctuation in the results of \ac{3dcae} is quite high.
We can further observe that the effect of the \ac{mnf} on the performance is rather weak.
The \ac{3dfeats} features lead to a slightly better performance boost, but still not enough to beat SSC.
Nevertheless, \ac{3dcae}, MS-2 and our approach achieve considerable scores.

\subsubsection{\acl{sa}}
On the \ac{sa} dataset, SSC yields a score on the raw data that is slightly better than ours.
The \ac{mnf} transform boosts the performance significantly and lets our approach beat SSC substantially.
Even the mean score is higher.
However, the standard deviation is very high in case of our framework, which indicates that we have a very high dependency on the initialization because of the highly non-convex problem.
Notably, the dark green and the mint green segment in the upper left part of the \ac{sa} ground truth (cf. \cref{img:sa-raw-gt}) describe the classes \emph{vineyard untrained} and \emph{grapes untrained}.
Both classes show a very high spectral similarity.
All competing approaches have problems to separate these two classes, resulting in segments that are usually completely mixed up or only classified as the same segment.
Merely our approach separates these two segments very well for both the raw (cf. \cref{img:sa-raw-an2norm-seg}) and the \ac{mnf}-transformed data (cf. \cref{img:sa-mnf-an2norm-seg}).

\subsubsection{Effect of MNF}
The application of the \ac{mnf} shows clearly a positive effect on the performance and gives a significant performance boost.
In case of \ac{ksc}, the \ac{mnf} transform is responsible that the methods can actually find a useful segmentation.
Notably, on \ac{pu} the \ac{mnf} causes a performance drop for $k$-means, \ac{gmm}, \ac{bgm}, \ac{3dcae} and MS-2 in terms of \ac{oa} but leads to an increase of the \ac{aa} scores, except for \ac{3dcae}.
This hints that the \ac{mnf} transform indeed helps to disentangle the data and allows to find also smaller spectral clusters and increase the accuracy on the corresponding segments.
A downside of the \ac{mnf} is that it tends to increase the standard deviation of the scores.
A possible reason is that, as already described in \cref{subsec:preprocessing}, the \ac{mnf} is not informed about the segment boundaries and therefore flattens differences in the boundary spectra.
This results in some spectra that move towards neighboring clusters and increase the potential of misclassifications of these particular spectra.
Depending on the initialization, this effect can lead to very different results.

\subsubsection{Effect of FL}
There is no clear tendency of the effect of the feature learning visible.
While in case of \ac{ip} and \ac{sa} the performance significantly drops in comparison to the raw and \ac{mnf} data in terms of \ac{oa}, the learned features yield a performance boost on \ac{pu} for almost all methods.
When considering the \ac{aa} score, however, we observe a drop in performance also on \ac{pu}.
The reason could be that the network used to learn the features applies 3D convolutions, meaning that in the feature vector of a pixel the information of the pixel plus its neighborhood is stored.
This helps in case of large segments to spectrally separate them from the other clusters, but, as there are many thin structures present in \ac{pu}, leads also to the effect that learned features of pixels lying in thin structures contain too much information about their spatial neighborhood and, therefore, are more similar to the neighboring features leading to misclassifications.
Hence, large segments can benefit from the learned features pushing the \ac{oa} score, while the accuracy of smaller segments drops, which lets the \ac{aa} score decrease.

\section{Conclusion}\label{sec:conclusion}
\noindent
In this article, we presented a new unsupervised segmentation framework for hyperspectral images based on the \ac{ms} segmentation functional equipped with a novel robust distribution-dependent indicator function that was designed to handle the spectral variability of \ac{hsi} data.
The method has proven to be suitable for the task at hand by achieving considerable scores on all four test datasets.
In particular, it beat the competitors on three of the four datasets and yields a competitive result on the fourth one.
The method was even able to separate clusters with a high spectral similarity and could handle the spectral variability adequately.
Moreover, we proposed an efficient fixed point iteration scheme to optimize the objective function with respect to the parameters for which, to the best of our knowledge, no closed form solution is available.

As the next step, we will focus on the noise estimation scheme for the \ac{mnf} transform, which currently is rather simplistic.
We believe that by replacing it by a scheme that uses assumptions that are more realistic and suitable for \ac{hsi} data, the segmentation results can be further improved.
Moreover, the estimation of the noise covariance matrix was done globally.
We plan to apply the \ac{mnf} transform only locally on adaptive neighborhoods with the goal to preserve the segment boundaries, similar to the approach in \cite{GuBa18}.
These adaptive neighborhoods can be either obtained by applying a superpixel segmentation method or by running our segmentation framework once on the data and rerunning it after applying the \ac{mnf} transform on each of the segments found in the first run separately.

\section*{Acknowledgments}
\noindent
This work was funded by the Deutsche Forschungsgemeinschaft (DFG, German Research Foundation) -- 333849990/GRK2379 (IRTG Modern Inverse Problems).

The research of Chandrajit Bajaj was additionally supported in part by the Peter O'Donnell Foundation.
\bibliography{references_resubmission}
\bibliographystyle{IEEEtran}

\begin{IEEEbiographynophoto}{Jan-Christopher Cohrs}
	is currently a Ph.D. student at AICES, RWTH Aachen University, Germany. He is supervised by Prof. Benjamin Berkels. His research interests include variational and unsupervised machine learning techniques for hyperspectral imaging.
\end{IEEEbiographynophoto}
\begin{IEEEbiographynophoto}{Chandrajit Bajaj}
	is a Professor of Computer Sciences at the University of Texas at Austin and the director of the Center for Computational Visualization, at the Oden Institute for Computational and Engineering Sciences. Bajaj holds the Computational Applied Mathematics Chair in Visualization. He is also an affiliate faculty member of Mathematics, Electrical Engineering, Bio-Medical Engineering, Neuroscience. He is a fellow of the American Association for the Advancement of Science (AAAS), a fellow of the Association of Computing Machinery (ACM), a fellow of the Institute of Electrical and Electronics Engineers (IEEE), and a fellow of the Society of Industrial and Applied Mathematics (SIAM).
\end{IEEEbiographynophoto}
\begin{IEEEbiographynophoto}{Benjamin Berkels}
	is a Juniorprofessor for Mathematical Image and Signal Processing at AICES, RWTH Aachen University, Germany. His research interests include variational and joint methods for image registration and segmentation, and medical image processing.
\end{IEEEbiographynophoto}
\end{document}